\documentclass[10pt,twocolumn,letterpaper]{article}

\usepackage[pagenumbers]{cvpr} 

\usepackage{url}

\usepackage{times}
\usepackage{epsfig}
\usepackage{graphicx}
\usepackage{amsmath}
\usepackage{amssymb}
\usepackage{makecell} 
\usepackage{booktabs,multirow,array} %
\usepackage{colortbl}
\usepackage{rotating}
\usepackage{pifont}
\usepackage{utfsym}
\usepackage{wrapfig}
\usepackage{tabularx}
\usepackage{xspace}
\usepackage[utf8]{inputenc} 
\usepackage[T1]{fontenc}    
\usepackage{amsfonts}       
\usepackage{mathtools} 
\usepackage{nicefrac}       
\usepackage{microtype}      
\usepackage[font={small}]{caption}
\usepackage[hang,flushmargin]{footmisc} 

\usepackage{makecell}
\usepackage[font={small}]{caption}

\usepackage{amsmath}

\usepackage{colortbl} 
\usepackage{amssymb} 
\usepackage{graphicx}
\usepackage{pifont}

\usepackage{caption}

\definecolor{antiquefuchsia}{rgb}{0.57, 0.36, 0.51}
\definecolor{cite}{RGB}{65,105,225}

\definecolor{colorbest}{RGB}{255,179,179}
\definecolor{colorsecond}{RGB}{255,217,179}
\definecolor{colorthird}{RGB}{255,255,179}

\definecolor{cvprblue}{rgb}{0.21,0.49,0.74}
\usepackage[pagebackref,breaklinks,colorlinks,citecolor=cvprblue]{hyperref}

\DeclareRobustCommand{\legendsquare}[1]{%
  \textcolor{#1}{\rule{2ex}{2ex}}%
}
\definecolor{colorbestLA}{RGB}{249,188,187}
\definecolor{colorbestLB}{RGB}{233,181,148}
\definecolor{colorbestLC}{RGB}{230,210,160}
\definecolor{colorbestLD}{RGB}{200,205,180}

\newcommand{\colorbestLD}[0]{\cellcolor{colorbestLD}}

\definecolor{colorbestNA}{RGB}{249,188,187}
\definecolor{colorbestNB}{RGB}{233,181,148}
\definecolor{colorbestNC}{RGB}{230,210,160}
\definecolor{colorbestND}{RGB}{200,205,180}
\definecolor{colorbestNE}{RGB}{192,210,227}

\newcommand{\colorbestNC}[0]{\cellcolor{colorbestNC}}

\newcommand{\colorbestNE}[0]{\cellcolor{colorbestNE}}

\PassOptionsToPackage{dvipsnames,table}{xcolor}
\usepackage{xcolor}
\usepackage{graphicx}
\usepackage{booktabs}
\usepackage{amsmath}
\usepackage{multirow}
\usepackage{makecell}
\usepackage{csquotes}
\usepackage{wrapfig}
\usepackage{diagbox}

\usepackage[utf8]{inputenc}
\usepackage{color,soul}
\usepackage[T1]{fontenc}

\definecolor{trashbin_blue}{RGB}{23,211,253}
\definecolor{hydrant_green}{RGB}{37,253,53}
\definecolor{bench_orange}{RGB}{255,195,45}

\setul{0.5ex}{0.3ex}
\setulcolor{trashbin_blue}

\usepackage{CJKutf8}

\definecolor{mygray}{RGB}{230,230,230}

\definecolor{clipcolor}{gray}{1.0}

\definecolor{openclipcolor}{gray}{1.0}

\definecolor{evaclipcolor}{gray}{1.0}

\definecolor{siglipcolor}{gray}{1.0}

\definecolor{llavacolor}{gray}{1.0}

\definecolor{sizescolor}{RGB}{194,215,255}

\definecolor{sizemcolor}{RGB}{165,202,255}

\definecolor{resscolor}{RGB}{194,255,191}

\definecolor{resmcolor}{RGB}{152,243,165}

\usepackage[most]{tcolorbox}

\colorlet{mapcolor}{ForestGreen}
\colorlet{langcolor}{RoyalBlue}
\colorlet{visioncolor}{WildStrawberry}
\colorlet{colabcolor}{YellowOrange}

\newtcbox{\cvtag}{enhanced,nobeforeafter,tcbox raise base,boxrule=0.4pt,top=0mm,bottom=0mm,
  right=0mm,left=4mm,arc=2pt,boxsep=2pt,before upper={\vphantom{dlg}},
colframe=visioncolor!50!black,coltext=visioncolor!25!black,colback=visioncolor!10!white,
  overlay={\begin{tcbclipinterior}\fill[visioncolor!80] (frame.south west)
    rectangle node[text=white,font=\sffamily\bfseries\tiny,rotate=90] {CV} ([xshift=4mm]frame.north west);\end{tcbclipinterior}}}

\newtcbox{\sharetag}{enhanced,nobeforeafter,tcbox raise base,boxrule=0.4pt,top=0mm,bottom=0mm,
  right=0mm,left=4mm,arc=2pt,boxsep=2pt,before upper={\vphantom{dlg}},
colframe=langcolor!50!black,coltext=langcolor!25!black,colback=langcolor!10!white,
  overlay={\begin{tcbclipinterior}\fill[langcolor!80] (frame.south west)
    rectangle node[text=white,font=\sffamily\bfseries\tiny,rotate=90] {TRA} ([xshift=4mm]frame.north west);\end{tcbclipinterior}}}

\newtcbox{\navitag}{enhanced,nobeforeafter,tcbox raise base,boxrule=0.4pt,top=0mm,bottom=0mm,
  right=0mm,left=4mm,arc=2pt,boxsep=2pt,before upper={\vphantom{dlg}},
colframe=colabcolor!50!black,coltext=colabcolor!25!black,colback=colabcolor!10!white,
  overlay={\begin{tcbclipinterior}\fill[colabcolor!80] (frame.south west)
    rectangle node[text=white,font=\sffamily\bfseries\tiny,rotate=90] {NAV} ([xshift=4mm]frame.north west);\end{tcbclipinterior}}}

\newtcbox{\loctag}{enhanced,nobeforeafter,tcbox raise base,boxrule=0.4pt,top=0mm,bottom=0mm,
  right=0mm,left=4mm,arc=2pt,boxsep=2pt,before upper={\vphantom{dlg}},
colframe=mapcolor!50!black,coltext=mapcolor!25!black,colback=mapcolor!10!white,
    overlay={\begin{tcbclipinterior}\fill[mapcolor!80] (frame.south west)
        rectangle node[text=white,font=\sffamily\bfseries\tiny,rotate=90] {LOC} ([xshift=4mm]frame.north west);\end{tcbclipinterior}}}

\newcommand{\shareul}[1]{\setulcolor{langcolor}\ul{#1}}
\newcommand{\navul}[1]{\setulcolor{colabcolor}\ul{#1}}
\newcommand{\locul}[1]{\setulcolor{mapcolor}\ul{#1}}

\tcbuselibrary{skins}

\AtBeginEnvironment{tcolorbox}{\footnotesize}

\usepackage[noorphans,vskip=1em,leftmargin=1em]{quoting}

\title{

Let Humanoids Hike!
Integrative Skill Development on Complex Trails
\\[-5mm]
}
\author{
Kwan-Yee Lin \qquad Stella X. Yu\\
University of Michigan\\
{\small\tt \{junyilin,\,stellayu\}@umich.edu}
}

\definecolor{lightgreen}{rgb}{0.6, 0.9, 0.6}
\definecolor{forest}{rgb}{0.0, 0.3, 0.0}

\begin{document}

\twocolumn[{
\renewcommand\twocolumn[1][]{#1}
\vspace{-9.5mm}
\maketitle%
\vspace{-13.5mm}
\begin{center}%
\includegraphics[width=0.99\textwidth]
 {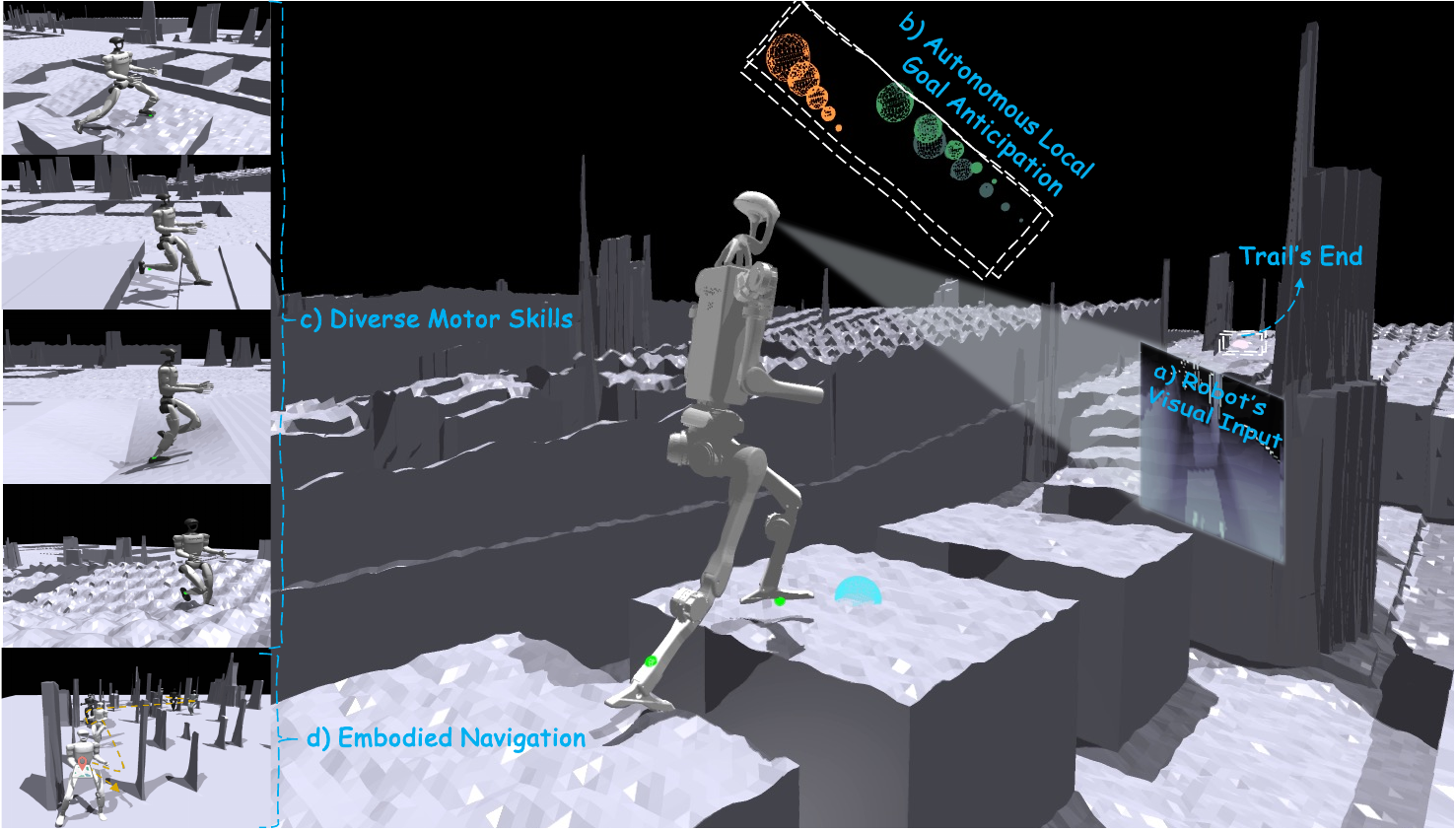}

\vspace{-8pt}%
\captionof{figure}{
\textbf{We propose training humanoids to hike complex trails, driving integrative skill development across visual perception, decision-making, and motor execution.}
{\bf Center:} The humanoid robot (H1) {\bf a)} equipped with vision, learns to {\bf b)} anticipate near-future local goals to guide locomotion along the trail with self-autonomy.
Bubble size (large $\rightarrow$ small) indicates anticipated goal direction; color shows temporal order (\textcolor{orange}{orange} $\rightarrow$ \textcolor{lightgreen}{green} $\rightarrow$ \textcolor{forest}{forest}).
{\bf Left:} Our LEGO-H framework is universal to different humanoid robots (\textit{e.g.,} G1, a smaller robot) to adaptively {\bf c)} emerge diverse motor skills, and {\bf d)} develop embodied path exploration strategies to hike on trails with varied terrains and obstacles.
Project page: \url{LEGO-H-HumanoidRobotHiking.github.io}.
}
\label{fig:teaser}
\end{center}
}
]

\begin{abstract}

Hiking on complex trails demands balance, agility, and adaptive decision-making over unpredictable terrain. Current humanoid research remains fragmented and inadequate for hiking: locomotion focuses on motor skills without long-term goals or situational awareness, while semantic navigation overlooks real-world embodiment and local terrain variability.  We propose training humanoids to hike on complex trails, driving integrative skill development across visual perception, decision making, and motor execution.

We develop a learning framework, LEGO-H, that enables a vision-equipped humanoid robot to hike complex trails autonomously.  We introduce two technical innovations:
{\bf 1)} A temporal vision transformer variant - tailored into Hierarchical Reinforcement Learning framework - anticipates future local goals to guide movement, seamlessly integrating locomotion with goal-directed navigation.
{\bf 2)} Latent representations of joint movement patterns, combined with hierarchical metric learning - enhance Privileged Learning scheme - enable smooth policy transfer from privileged training to onboard execution.
These components allow LEGO-H to handle diverse physical and environmental challenges without relying on predefined motion patterns. Experiments across varied simulated trails and robot morphologies highlight LEGO-H’s versatility and robustness, positioning hiking as a compelling testbed for embodied autonomy and LEGO-H as a baseline for future humanoid development.

\end{abstract}

\section{Introduction}
\label{sec:intro}

Hiking~\cite{nordbo2015hiking,mitten2018hiking} challenges humans to master diverse motor skills and adapt to complex, and unpredictable terrain -- such as steep slopes, wide ditches, tangled roots, and sudden elevation changes (Fig.~\ref{fig:teaser}). It demands continuous balance, agility, and real-time decision-making, making it an ideal testbed for advancing humanoid autonomy and the integration of vision, planning, and motor control. Hiking-capable robots could explore remote areas, assist in rescue missions, and guide individuals along rugged paths.

Hiking poses challenges beyond traditional navigation, blind locomotion, or single motor pattern learning. To succeed, humanoid robots must master three core capabilities: 
{\bf 1) Locomotion versatility} – The ability to handle mixed terrains like dirt, rocks, stairs, and streams, adapting dynamically with skills like jumping and leaping while maintaining balance. {\bf 2) Perceptual awareness} - The ability to sense and respond to complex 3D environments, such as stepping over logs or navigating around trees. {\bf 3) Body awareness} – The ability to adjust in real time to local obstacles, terrain changes, and body states by coordinating vision and motor control for adaptive foot placement and movement.

\begin{figure}[t!]
    \centering
    \includegraphics[width=0.9\linewidth]{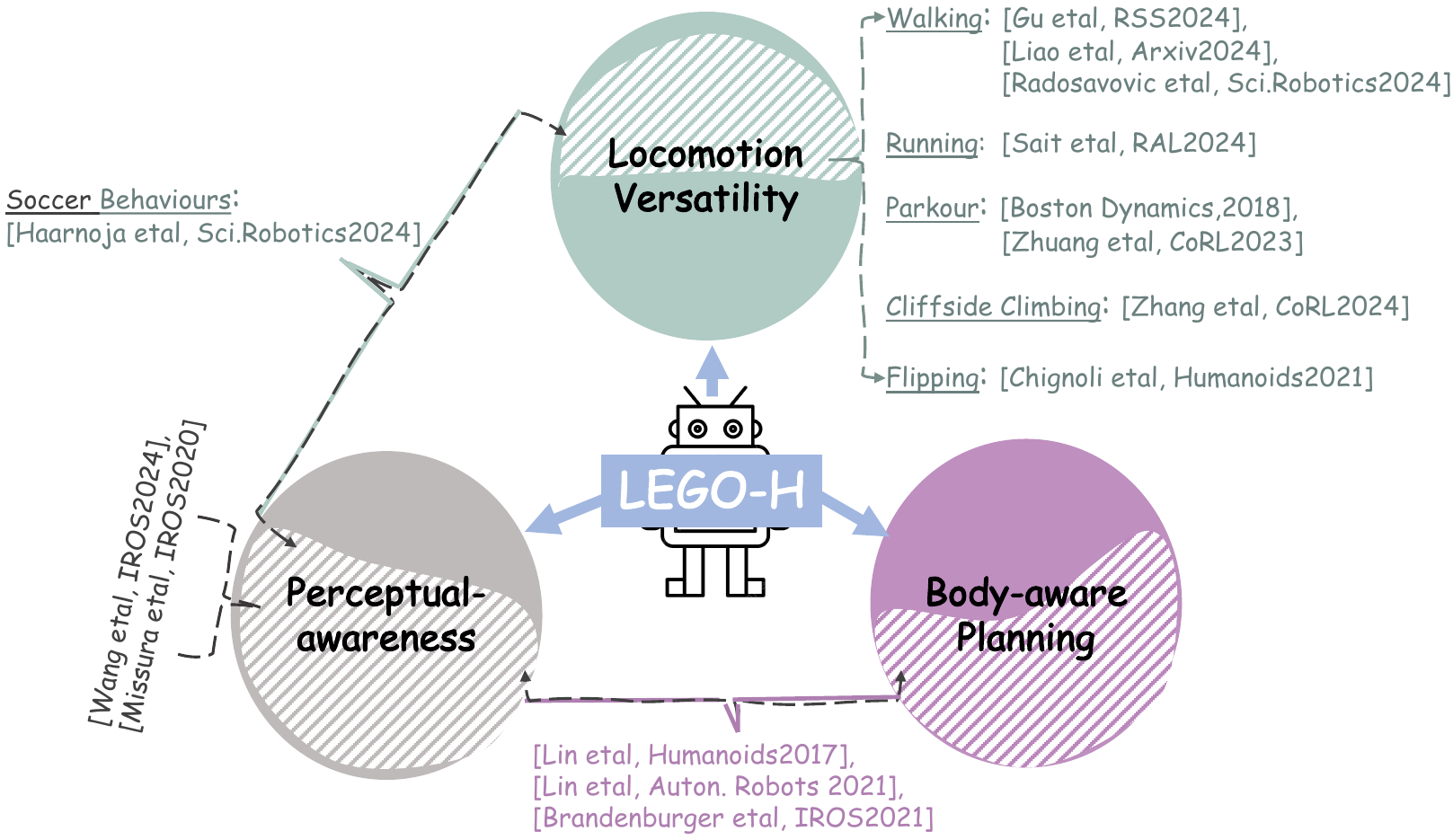}
 \vspace{-0.1in}
 \caption{\textbf{
Hiking requires locomotion versatility, perceptual awareness, and body-aware planning - integrated for the first time in our approach.} 
Prior work considers only subsets of these capabilities (hatched patterns), whereas LEGO-H unifies all three within a single learning framework to enable embodied autonomy.}
    \label{fig:method-diff}
\end{figure}

Current humanoids struggle to meet these demands due to the lack of a unified framework that integrates low-level motor skills with high-level navigation (Fig~\ref{fig:method-diff}). 
{\bf1) Locomotion methods lack adaptability to terrain variation}.  They treat terrain as a fixed, homogeneous, and passive background, focusing narrowly on walking~\cite{DBLP:journals/scirobotics/RadosavovicXZDMS24,DBLP:journals/corr/abs-2303-03381}, quasi-periodic motion patterns~\cite{DBLP:conf/iclr/LiSHK24}, or mimicry~\cite{2021-TOG-AMP}. Advanced frameworks for complex skills like parkour~\cite{DBLP:journals/corr/abs-2406-10759,boston}, often depend heavily on user commands or engineered behaviors. Such isolated
training paradigms and abstraction overlook the embodied interaction essential for real-world locomotion, limiting generalization beyond curated environments.
{\bf 2) Navigation methods struggle with real-time adaptability.} Traditional research efforts rely on scene mapping~\cite{5971330} or rigid world geometry ~\cite{DBLP:conf/iros/MissuraRB20}. 
While LLMs and VLMs can plan behaviors and correct execution failures from textual instructions~\cite{DBLP:journals/corr/abs-2408-08282}, they often lack the physical grounding needed for real-world adaptability. A robot may know it needs to {\it step over the log}, but without real-time perception and fine-grained motor control, it cannot adjust mid-swing if the log shifts or the ground gives way. Reflexive foot placement on uneven terrain demands fast, sensor-driven adaptation - not just faster planning - which symbolic planners struggle to provide. Bridging motor skills and navigation remains challenging due to their inherently different response levels (fast, reactive control vs. slower, deliberative planning) requiring tight coordination for context-sensitive execution in complex environments.

We introduce LEGO-H, a perceptual-aware, end-to-end, embodied learning framework for acquiring situational visual-motor skills and path exploration strategies that enable humanoids to traverse complex trails autonomously (Fig.~\ref{fig:teaser}). It unifies navigation and locomotion by advancing \underline{H}ierarchical \underline{R}einforcement \underline{L}earning (HRL) and enhancing \underline{P}rivileged \underline{L}earning (PL) for effective skill development.

{\textbf{Our first technical contribution is task-grounded HRL for situational visual-motor control, reformulating navigation as a sequential local goal anticipation problem to guide locomotion policy learning}}. While HRL can unify navigation and locomotion via multi-level abstraction, existing methods often
oversimplify environments~\cite{DBLP:conf/icra/GarimortHB11}, or restrict low-level control to basic skills like walking~\cite{DBLP:journals/firai/AhnJBS21}, limiting adaptability. We address this caveat by proposing TC-ViT,
a temporal vision transformer variant tailored for HRL that combines tokenization with embodied reinforcement learning.  Instead of treating the navigation target as a static token, TC-ViT models {\bf 1)} navigation goals and {\bf 2)} temporal-spatial relations, considering the robot's past, present, and future states for sequential anticipation. The locomotion policy network then integrates these latent features with proprioceptive inputs and partial anticipated navigation goals to produce motor actions, enabling tight coordination between perception and control for navigating complex, dynamic trails.

{\textbf{Our second technical contribution is enhanced PL that distills diverse motor skills while preserving action rationality.}}  In PL, a teacher policy leverages privileged signals such as known foothold locations to develop diverse, optimal behaviors efficiently and safely. A student policy then learns to replicate these behaviors using only proprioception and onboard perception, enabling deployment in unstructured environments without privileged information. It improves skill acquisition but complicates action learning when integrating visual inputs, increasing the risk of errors and damage from unexpected actions. Existing distillation approaches supervise global behaviors~\cite{tessler2023calm} or per-joint accuracy~\cite{kumar2021rma}, often ignoring inter-joint dependencies. We address this by proposing a Hierarchical Latent Matching (HLM) metric 
that distills policy based on action rationality.
HLM utilizes structured latent representations and masked reconstruction via VAEs~\cite{DBLP:journals/corr/KingmaW13} to enforce relational consistency across joints. This task-agnostic HLM loss set improves policy learning across motor tasks. Crucially, the latent prior is derived from oracle policy, {\it not} human demonstrations, allowing robot to learn self-reliant behaviors suited to its own morphology.

To summarize, our work makes three key contributions:
{\bf 1)} We propose hiking as a testbed for integrative skill development in humanoid robots.
{\bf 2)} We introduce LEGO-H, a learning framework for autonomous humanoid hiking.
{\bf 3)} We demonstrate LEGO-H’s robustness and versatility across diverse simulated trails and humanoid morphologies, establishing hiking as a compelling testbed for embodied autonomy and LEGO-H as a baseline for future humanoid research.

\section{Related Work}
\label{sec:relatedwork}
\noindent\textbf{{Humanoid locomotion.}}~Existing approaches to low-level motor skill learning typically simplify environmental interactions, abstracting terrains into static patterns at a {\textit{momentary}} scale, which neglects occlusions caused by obstacles or dynamic environmental disruptions. Research in this domain has primarily focused on learning specific locomotion skills such as walking~\cite{DBLP:journals/scirobotics/RadosavovicXZDMS24,DBLP:journals/corr/abs-2303-03381,DBLP:journals/corr/abs-2402-16796,DBLP:journals/corr/abs-2407-21781,gu2024advancing}, running~\cite{DBLP:journals/ral/SovuklukEO24,DBLP:journals/firai/SmaldoneSLO22}, and soccer-playing behaviors~\cite{DBLP:journals/scirobotics/HaarnojaMLHTHWTSHBHBHTSBCSG24}. These approaches often rely on highly engineered designs optimized for specific lower-body tasks. Other works employ imitation learning~\cite{DBLP:conf/iclr/LiSHK24,2021-TOG-AMP,2022-TOG-ASE,DBLP:conf/iclr/0002CMWHKX24,tessler2024maskedmimic} to generate human-like behaviors from large-scale motion datasets, but this comes at the cost of reduced embodiment. Some frameworks attempt to push the boundaries of robotic motor skills by exploring tasks like parkour~\cite{boston,DBLP:journals/corr/abs-2406-10759}, acrobatic flipping~\cite{DBLP:conf/humanoids/ChignoliKSK21}, or cliffside climbing~\cite{DBLP:journals/corr/abs-2406-06005}. While impressive, these methods are often bogged down by complex engineering, reliance on user commands for motion planning, or lack of perceptual awareness.

\noindent\textbf{{Humanoid navigation.}}~Research on this direction often struggles to address {\textit{real-time}} environmental constraints while accounting for the unique mechanisms and actions of humanoid robots. These limitations frequently lead to suboptimal navigation plans in complex terrains. Conventional methods typically rely on scene mapping~\cite{5971330,DBLP:journals/corr/abs-2210-03280} or structured world assumptions~\cite{DBLP:conf/iros/MissuraRB20}, which restrict adaptability in dynamic and unstructured environments. Contact-aware approaches~\cite{DBLP:conf/humanoids/LinB17,DBLP:journals/arobots/LinB21} attempt to bridge robot configurations with environmental constraints, but they often depend on pre-generated trajectories, limiting responsiveness. Similarly, mapless methods~\cite{DBLP:conf/iros/BrandenburgerRB21} leverage visual inputs for navigation but are typically constrained to basic locomotion capabilities such as walking. Recent advancements in large language and vision-language models have shown potential for complex high-level planning~\cite{DBLP:journals/corr/abs-2408-08282}, yet remain uncoupled from motor control systems, failing to achieve autonomous perceptual awareness and last-step feasibility required for navigating diverse, fine-grained environments, like hiking.

\noindent\textbf{{Joint learning of navigation and locomotion.}}~Integrating navigation and locomotion into a unified framework remains a significant challenge. In the realm of wheeled-legged and quadruped robots, several studies~\cite{DBLP:journals/scirobotics/LeeBRWMH24,DBLP:conf/iclr/YangZHXW22,DBLP:journals/ral/HoellerWFH21,DBLP:conf/iros/RudinHB022} have explored paradigms that unify local navigation and locomotion. While these approaches provide valuable insights, tailoring them to humanoid robots as a baseline for hiking tasks reveals several critical gaps. First, humanoid robots possess significantly more degrees of freedom (DoF) than quadrupeds or wheeled-legged robots, complicating the development of stable locomotion policies. Achieving balance across diverse lower-body motor skills ({\textit{e.g.,}} walking, jumping, and leaping~\textit{etc.}) within a single framework remains an open problem. Second, the greater body height of humanoid robots introduces challenges in visual perception, expanding their field of view and capturing a broader range of distances. This increased perceptual complexity exacerbates the misalignment between environmental sensing and physical contact, further complicating decision-making, navigation, and motor execution processes.

{\textit{See Appendix for more related works on HRL and PL}}.
\section{LEGO-H for Integrative Skill Learning}

\subsection{Task Definition}\label{task_def}
Drawing from human hiking paradigm~\cite{allt}, we consider a humanoid robot equipped with vision and GPS. A hiking trail is specified by start and end points $(P_A, P_B)$ in GPS, optionally with $M$ intermediate waypoints along the trail. We define the basic task of {\it humanoid hiking} as follows: {\it traversing a trail to reach the trail's end $P_B$ with safety, efficiency, and all-level autonomy}.

The robot receives the following inputs:
\begin{enumerate}[leftmargin=*,topsep=1pt,itemsep=1pt]
\item GPS-based 2D vector $D_{rb}$ from robot’s current projected 2D root position $P_R[:2]$ to end $P_B[:2]$, which may not be visible from start $P_A$. This vector provides the distance and direction of the endpoint relative to the robot. 

\item GPS-based 2D vectors $\{D_{rm}\}_{m=1}^M$ from $P_R$ to $M$ optional intermediate waypoints. We use $M\!=\!1$ to study the basic trail structure and disambiguate forks.  These points provide guidance but need not be strictly followed.
\item The onboard proprioceptive input $\mathcal{X}_{pro}$, like joint velocities and angles, reflects the robot’s internal physical state.

\item $K$ forward-facing depth images $\{C_k\}_{k=1}^K$ from a head-mounted camera. Unlike prior quadruped approaches~\cite{DBLP:conf/iros/RudinHB022,DBLP:journals/scirobotics/LeeBRWMH24} assuming full local 3D information, our setup limits vision to a frontal field, making perceptual-motor learning more realistic and challenging. Humanoids, being taller than quadrupeds, see farther - enabling look-ahead planning but complicating near-term action learning.
\end{enumerate}
For ideal hiking, whole-body control would allow coordinated use of arms and legs to maintain balance and support denser contact points with trails. However, as a baseline prototype for this new task -- and noting that many trails can still be traversed with leg movement alone -- this study simplifies the task by freezing humanoid's upper-body pose, focusing on lower-body functionality.

\subsection{LEGO-H System Overview}\label{method-overview}
\begin{figure}[t!]
    \centering    \includegraphics[width=1\linewidth]{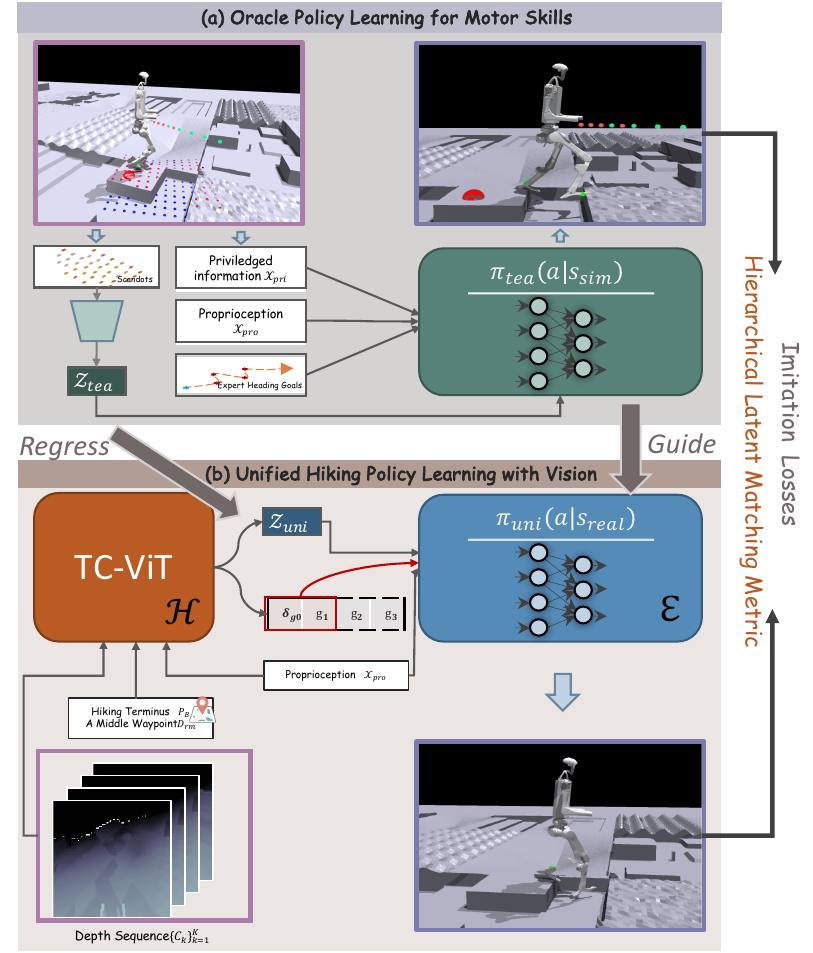}
    \vspace{-0.2in}
    \caption{\textbf{LEGO-H framework overview.} 
    LEGO-H equips humanoid robots with adaptive hiking skills by integrating navigation $\mathcal{H}$ and locomotion $\mathcal{E}$ in a unified, end-to-end learning framework ($b$). To foster the versatility of motor skills, we train the unified policy via privileged learning from the oracle policy ($a$).}
    \label{fig:method-overview}
\end{figure}

In our setup, the robot is only given the relative position of the endpoint. Thus, it must {\textit{autonomously}} determine how to traverse unknown, but locally observable trail with various terrain changes to reach the destination {\textit{safely}}. From a framework perspective, a humanoid system must fulfill two core requisites to succeed: {\bf1) learn embodied path exploration that is both target-driven and locally adaptive} -- the robot must autonomously assess and adapt its local path based on immediate sensory
observations and current executable motor skills, while maintaining alignment with the overall
goal;  {\bf2) enable emergent, context-aware, and safe motor execution} -- the robot must learn a diverse set of motor skills and execute actions that are not only safe for its body but also feasible under local environmental constraints, like clearance and terrain support. To this end, we propose an end-to-end, embodied learning framework, LEGO-H (Fig.~\ref{fig:method-overview}), short for {\it {\underline{Le}t Humanoids \underline{Go} \underline{H}iking}}.

To fulfill the first requisite, LEGO-H employs two levels of modules within a unified policy learning pipeline (Fig.~\ref{fig:method-overview}b), combing a high-level navigation module ($\mathcal{H}$) that encodes trail's latent representation and anticipates local goals,  with a low-level motor skill module ($\mathcal{E}$) that learns reactive motor policy in real time. Specifically:
\begin{enumerate}[leftmargin=*,topsep=1pt,itemsep=1pt]
\item 
The high-level navigation module $\mathcal{H}$, implemented via TC-ViT (Sec.~\ref{TC-ViTs}), 
acts as a {\it trail scout}, looking ahead and proposing local directions based on visual cues, global goal, and motor execution.  It
receives the state $\mathbf{s}_{real}$ (depth images $\{C_{k}\}^K_{k=1}$, proprioception $\mathcal{X}_{pro}$, endpoint $P_B$, and one middle waypoint $M$), generates a latent trail representation $\mathbf{z}_{uni}$, anticipates a sequence of $N$ {\textit{future}} local navigation goals $G\!=\!\{g_{n}\}^{N}_{n=1}$, and calculates a goal residual $\delta{g_0}$ capturing the execution mismatch from the {\textit{previous step}}. Each $g_n \in [0, 2\pi]$ represents a goal direction as a yaw angle relative to the robot’s root.  
\item 
Then, the latent trail representation $\mathbf{z}_{uni}$, proprioception $\mathcal{X}_{pro}$, residual $\delta_{g_0}$, and the {\textit{next}} anticipated goal $g_1$, flow to the low-level motor skill module $\mathcal{E}$ to guide {\textit{softly}}. $\mathcal{E}$
plays the role of an agile {\it trail runner}, 
reacting in real time to proprioceptive feedback and terrain conditions to decide how best to execute each step.
It predicts an executable action $\mathbf{a}_t$. Rather than strictly tracking the sequence of local goals from $\mathcal{H}$, $\mathcal{E}$ adapts to local terrain and robot state to safely progress toward the endpoint.

\end{enumerate}
By seamlessly leveraging visual and proprioceptive feedback within an RL framework, this unified pipeline reflects HRL's abstraction, where local goal anticipation and reactive control jointly enable the robot to autonomously adapt local paths within traversable regions, avoiding entrapment and collisions in challenging trail
terrains, while maintaining steady progress toward the trail's end.

LEGO-H achieves the second requisite by enhancing privileged learning scheme wrt structural rationality of actions:

\begin{enumerate}[leftmargin=*,topsep=1pt,itemsep=1pt]
\item It first trains an oracle motor skill policy $\pi_{tea}({\bf a}|\mathbf{s}_{sim})$ (Sec.~\ref{method-p1}) with privileged information $\mathcal{X}_{pri}$ (\textit{e.g.,} terrain type, ground friction, precise state measurements) and expert navigation goals as inputs (Fig.~\ref{fig:method-overview}a). While vision is not used at this stage, scandots and $\mathcal{X}_{pri}$ provide clean, informative signals for high-quality skill acquisition. 
\item Then, in the unified pipeline training, the teacher policy is distilled into $\mathcal{E}$ to initialize it. Aside from basic imitation losses and rewards (Sec.~\ref{method-p2}), LEGO-H uses a Hierarchical Latent Matching metric (Sec.~\ref{hlm}) to learn the final policy $\pi_{uni}({\bf a}|\mathbf{s}_{real})$ that balances robustness and behavioral diversity across diverse trail terrains.
\end{enumerate}

\subsection{TC-ViT: Autonomous Local Goal Anticipation}\label{TC-ViTs}

\begin{figure}[t!]
    \centering
    \includegraphics[width=1\linewidth]{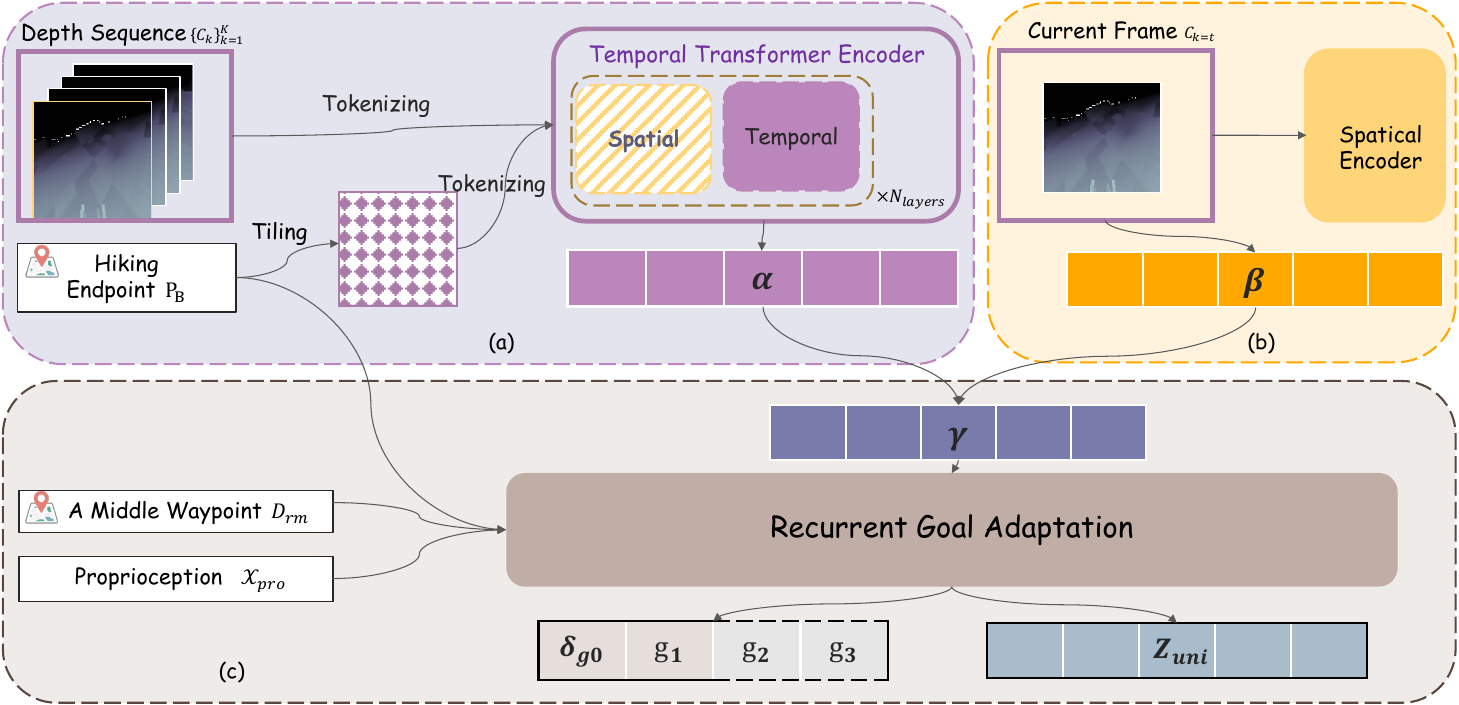}
\vspace{-0.2in}
    \caption{\textbf{TC-ViT Architecture.} Three key components: {\bf a)} a goal-orientated temporal transformer encoder for robots cognizing surroundings with the final goal; {\bf b)} a parallel process on the current depth frame for integrating spatially precise information to reflect the current state {\bf c)} a recurrent goal adaptation mechanism that integrates visual awareness, goal information, and proprioception.
    }

    \label{fig:tcvits}
\end{figure}

The navigation module $\mathcal{H}$ is implemented via TC-ViT, a variant of \underline{T}emporal Information \underline{C}onditioned \underline{Vi}sion \underline{T}ransformer.
It serves as a central mechanism to achieve unified policy learning with visual perception, by addressing four critical aspects to navigation module: $1)$ cognize surroundings with balance of short-time reactivity and final goal alignment, adapt anticipation of local goals to local terrain with $2)$ spatial precision and $3)$ embodied awareness, and $4)$ produces representations with synchronized perception and action (shown in Fig.~\ref{fig:tcvits}).

\noindent\textbf{1) Cognize surroundings with final goal.} A common strategy for environment perception assumes Markovian observations and processes adjacent depth images via methods like 3D modeling~\cite{DBLP:journals/scirobotics/LeeBRWMH24}/reconstruction~\cite{yang2023neural}, temporal features~\cite{DBLP:conf/icra/ChengSAP24}, or semantic traversability~\cite{mattamala24wild}. However, hiking poses two key challenges:
{\bf 1)} {\textit{Time scale}}: short-term dynamics and long-term environmental dependencies must be handled jointly.
{\bf 2)} {\textit{Specificity}}: Visual features must directly support execution of immediate next step while aligning with final goal.

Thus, a direct solution is to integrate local perception with a distant global goal $P_B$, where we employ a temporal vision transformer with goal conditioning (Fig.\ref{fig:tcvits}a), adapted from classic ViViT's encoder\cite{arnab2021vivit}. It captures the information with both spatial and long-range dependencies via processing 16-frame depth sequences (downsampled to 4) into spatio-temporal tokens (from $16\times16$ patches) using 6 transformer layers with spatial and temporal attention. 
The final goal $P_B$ is tiled as an additional $(1,\! H,\! W)$ channel ($H\!=\!W\!=\!128$) and fused at tokenization.  This early fusion ensures goal awareness is preserved throughout spatio-temporal reasoning, yielding more task-aligned predictions.  The encoder outputs a flattened feature vector $\boldsymbol{\alpha}(\{C_{k}\}^K_{k=1},P_{B})$. 

Intuitively, this part of TC-ViT serves as a trail scout with a map in hand: it interprets what’s immediately ahead through sequences of depth images, while constantly factoring in the direction of the final destination. Embedding the goal early - before visual abstraction — ensures the robot always ``looks'' with intent, allowing it to anticipate terrain-compatible moves that remain globally purposeful.

\noindent\textbf{2) Anticipate near-future goals with spatial precision.}
While aboves might be effective to support long-horizon goal prediction in coarse, body-agnostic navigation~\cite{DBLP:conf/corl/ShahSDSBHL23}, humanoid hiking demands fine-grained, multi-scale decision-making. On uneven trails with sudden obstacles (Fig.~\ref{fig:teaser}), precise foot placement and rapid balance adjustments are critical - capabilities that suffer as temporal transformers abstract away fine spatial structure critical for precise control.

The second component of TC-ViT (Fig.~\ref{fig:tcvits}b) thus introduces a parallel path focused on immediate perception. It processes the current depth image $C_{k=t}$ through a shallow CNN, producing high-resolution spatial features $\boldsymbol{\beta}(C_{k=t})$ that capture near-field terrain details. This branch omits goal conditioning, as its role is purely reactive.

The final representation $\boldsymbol{\gamma}$ combines long-range goal-informed context $\boldsymbol{\alpha}$ with fine-grained local perception $\boldsymbol{\beta}$ via feature concatenation followed by MLPs:
$\boldsymbol{\gamma} = \text{MLPs}(\text{concat}(\boldsymbol{\alpha}, \boldsymbol{\beta}))$.
Intuitively, this merges the foresight of a trail guide - who knows where the path leads - with the reflexes of a hiker watching their next step.

\begin{figure}[t!]
    \centering
    \vspace{-0.1in}
    \includegraphics[width=1\linewidth]{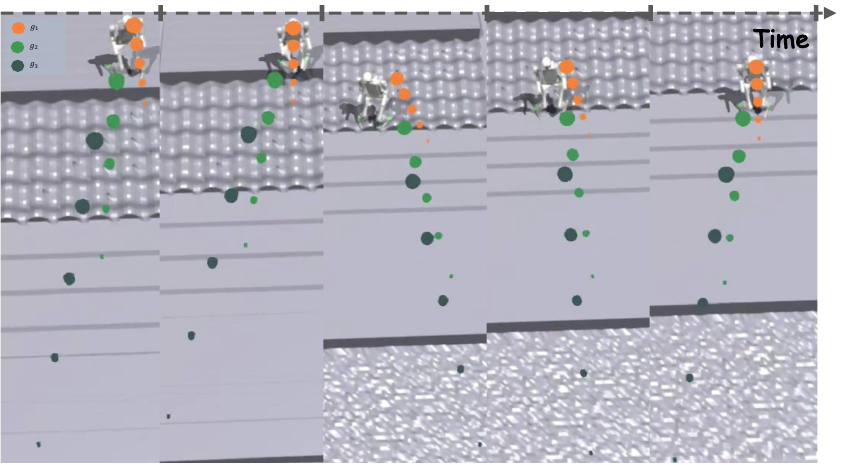}
     \vspace{-0.2in}
    \caption{\textbf{Dynamic adjustments of near goal anticipation.} Snapshots from left to right show a robot traversing mixed terrains along a trail. TC-ViT does not provide a fixed trajectory that locomotion module must rigidly follow. Instead, it predicts several near-future goals ($g_{1}, g_{2}, g_{3}$), which dynamically adapt to robot's current state, reflecting real-time adjustments to its navigation decisions. Bubble size (large$\rightarrow$ small) represents predicted local navigation direction.}

    \label{fig:goal-dynamic}
\end{figure}

\noindent\textbf{3) Adapt goals with embodied awareness.}
Beyond understanding environment, effective navigation must also account for how motor actions and body state affect outcomes. TC-ViT includes a third part - a recurrent goal adaptation mechanism (Fig.~\ref{fig:tcvits}c) - that fuses visual features, proprioception, and goal information to adaptively anticipate a sequence of local goals, and produce an embodied latent representation.

Specifically, inputs including the visual representation $\boldsymbol{\gamma}$, endpoint $P_B$, intermediate cue $\mathcal{D}_{rm}$, and proprioception $\mathcal{X}_{pro}$ are passed through a two-layer MLP and a GRU to model temporal dependencies:
$
\mathbf{z}_{uni}, \delta{g_0}, G\! =\! \text{GRU}(\text{MLPs}(\boldsymbol{\gamma}, P_B, \mathcal{D}_{rm}, \mathcal{X}_{pro})).
$
The resulting latent encodes perceptual context, physical embodiment. The residual correction $\delta g_0$ and near-future goals $G$  provide soft guidance for the locomotion module. Intuitively, this mechanism helps the robot learn not just what it sees, but how it moves through what it sees, adapting its local goals based on how past actions played out, and staying grounded in both vision and bodily awareness (as shown in Fig~\ref{fig:goal-dynamic}).

\noindent\textbf{4) Synchronize perception and control.}
Real-world systems operate at mismatched time scales, e.g.,  Unitree H1's depth sensing runs at $10\pm 2$ Hz with RealSense D435i, while control executes at 50 Hz on Jetson NX.  TC-ViT addresses this latency gap with two strategies.
\textit{1. Nearest-goal forwarding:} Only the immediate goal $g_1$ is passed to the locomotion module, ensuring timely response and reducing drift from delayed decisions (see Appendix). Intuitively, this reflects the idea that -- while multiple goals are anticipated, only the immediate one shapes action, as it reflects the step that matters right now.
\textit{2. Latent tiling:} The latent representation $\mathbf{z}_{uni}$ is tiled five times per control cycle to maintain a stable signal stream.
Together, these mechanisms bridge asynchronous modules and allow perception and action to stay in sync despite hardware-level delays.

\subsection{Oracle Policy Learning for Motor Skills}\label{method-p1}

Before unifying navigation and locomotion via TC-ViT, we pretrain an {\it oracle} locomotion policy (Fig.~\ref{fig:method-overview}a) to acquire diverse motor skills. The oracle takes as input proprioception $\mathcal{X}_{pro}$, current navigation goal, privileged state $\mathcal{X}_{pri}$, and latent terrain features $\mathbf{z}_{tea}$ from scandots $\mathcal{S} \in \mathbb{R}^{66 \times 2}$. To encourage upright locomotion with emergent motor behavior rather than pre-defined modes, rewards in three aspects are essential in this stage: {\bf 1)} direction-aligned velocity tracking $r_{\text{tracking}}$,
{\bf 2)} soft torso height constraint $r_{\text{base-height}}$,
{\bf 3)} foot airtime accumulation $r_{\text{air-time}}$.
See Appendix for details.

\subsection{Unified Hiking Policy Learning with Vision}\label{method-p2}

After training the oracle policy $\pi_{tea}(\mathbf{a}|\mathbf{s}_{sim})$, we distill it into a unified student policy $\pi_{uni}(\mathbf{a}|\mathbf{s}_{real})$ that jointly learns navigation and motor control from visual input (Fig.~\ref{fig:method-overview}b). Specifically, TC-ViT encodes depth sequences into  latent $\mathbf{z}_{uni}$ and predicts near-future goals. The tuple $(\mathbf{z}_{uni}, \delta_{g_0}, g_1)$ is passed to the locomotion module to compute $\pi_{uni}({\bf a}|{\bf s}_{real})$, which outputs current action ${\bf a}_t$. Both policies are implemented as MLPs.
Basic training losses here are RL rewards (see appendix) and reconstructions for imitation in goal, latent, and action levels from teacher stage:
\begin{align}
    \vspace{-0.2cm}
    \mathcal{L}_{im} &= w_1 \|\mathbf{z}_{tea} - \mathbf{z}_{uni}\|^2 + w_2 \, \text{SmoothL1}(\mathbf{G}_{tea}, \mathbf{G}_{uni}) \nonumber \\
    &\quad + w_3 \, \text{SmoothL1}(\mathbf{a}_{tea}, \mathbf{a}_{uni}).
\end{align}
The oracle acts as a mentor guiding student through complex terrain. By initializing $\pi_{uni}$ via imitation and optimizing it together with TC-ViT under RL framework, $\pi_{uni}$ learns to align vision, planning, and control into a cohesive behavior.

\subsection{Hierarchical Latent Matching Metric}\label{hlm}

Standard action imitation loss aggregates per-joint errors, overlooking joint coordination. Thus, we introduce {\underline{H}}ierarchical {\underline{L}atent {\underline{M}}atching (HLM) loss metric, which captures structural dependencies to bound the student's action space. We first train a masked VAE on oracle actions to learn a latent space that encodes joint coordination.  During distillation, student policy is guided to match this latent structure, promoting physically coherent and well-coordinated actions despite modality and representation gaps. Analogous to feature matching in image reconstruction, this method shifts imitation from pointwise joint matching to holistic joint pattern matching, treating the body as a coordinated system rather than a set of independent joints.

Specifically, during distillation, VAE is {\textit{iteratively}} trained on teacher actions with randomly masked joints, learning to reconstruct full actions from partial inputs, where:
\begin{align}
 \mathcal{L}_{rec} &= w_{4}\mathcal{L}_{KL}+ w_5\mathcal{L}_{self} + w_6\mathcal{L}_{mask}\\
\mathcal{L}_{KL} &= KL\left(q(\mathbf{z_{vae}}|\mathbf{a}_{\text{tea}}) \parallel \mathcal{N}(0, I)\right)
\label{kl-eq}\\
\mathcal{L}_{self} &= \text{SmoothL1}(\text{Dec}(\text{Enc}(\mathbf{a}_{\text{tea}})), \mathbf{a}_{tea})\\
\mathcal{L}_{mask} &= \text{SmoothL1}(\text{Dec}(\text{Enc}(\mathbf{a}_{\text{tmask}})), \mathbf{a}_{tea})
\end{align}
Here, $w_x$ are weighting terms, $\mathbf{z}_{vae}$ is latent vector, and $\mathbf{a}_{tmask}$ denotes masked teacher action. KL term follows VAE formulation~\cite{DBLP:journals/corr/KingmaW13}. To handle joint permutation invariance, we apply sine-cosine positional embeddings to each joint. The compact latent space, regularized by the Gaussian prior and enriched by masking, encourages learning of inter-joint dependencies and structural consistency, capturing coordination patterns aligned with the robot’s physical embodiment, rather than relying on human motion priors.

Once trained, the VAE encoder defines a structured feature space for comparing teacher and student actions. We utilize it to introduce a two-level HLM loss: full-feature alignment and masked-subset matching.

Concretely, for each student action $\mathbf{a}_{uni}$, we compute a cosine similarity loss with the teacher action: 
\begin{align}
\mathcal{L}_{ts} &=1-\text{cos\_sim}(\text{Enc}(\mathbf{a}_{tea}), \text{Enc}(\mathbf{a}_{uni})) \\
&= 1-\frac{\text{Enc}(\mathbf{a}_{tea}) \cdot \text{Enc}(\mathbf{a}_{uni})}{\|\text{Enc}(\mathbf{a}_{tea})\| \|\text{Enc}(\mathbf{a}_{uni})\|}
\end{align}
We further apply a triplet-style consistency loss using a randomly masked student action:
\begin{align}
    &\mathcal{L}_{trip} = c_{mt} (1-\text{cos\_sim}(\text{Enc}(\mathbf{a}_{tea}), \text{Enc}(\mathbf{a}_{umask})))\nonumber\\ 
    &+ c_{ms}(1-\text{cos\_sim}(\text{Enc}(\mathbf{a}_{uni}), \text{Enc}(\mathbf{a}_{umask})))
\end{align}
The combined hierarchical loss is: 
\begin{equation} 
\mathcal{L}{hie} = w_7 \mathcal{L}{ts} + w_8 \mathcal{L}_{trip} 
\end{equation}
See Appendix for hyperparameters. 
As shown in Tab.~\ref{tab:ablation_h1}, without HLM, student robots can complete the task but with frequent collisions and poor coordination. In contrast, HLM promotes robots to exhibit more refined, collision-free movements that align better with internal structural consistency.
\section{Experiments}
We evaluate effectiveness of LEGO-H across several dimensions. First, we conduct ablations (Sec.~\ref{ablation}) to assess individual components. Then, we analyze robot's emerged behaviors across different levels (Sec.~\ref{behaviors}). Finally, as a new task, we benchmark humanoid hiking in diverse simulated trail environments, covering LEGO-H, and other representative methodologies tailored to this task (Sec.~\ref{benchmark}). 
We detail experimental setup on robot configurations/models/evaluation metrics (Sec.~\ref{exp-settings}). {\textit{See Appendix for more details.}}

\subsection{Experimental Settings}\label{exp-settings}
\noindent\textbf{Robots.} We use Unitree H1~\cite{unitreeh1} and G1~\cite{unitreeg1} humanoids, chosen for their distinct differences in body scale and mechanism: H1, at adult size ($5.9$ ft/$47$kg), contrasts with kid-sized G1 ($4.26$ ft/$35$kg), with notable variations in torque density and morphology. These inherent differences impact key factors like visual perception range/motor stability/overall movement complexity even within identical trails.

\noindent\textbf{Implementations.} {\textit{Proprioception}} ($\mathcal{X}_{pro}\in \mathbb{R}^{40}$): covers lower-body joint positions, velocities, torso roll and pitch, foot contact indicators, and previous action $a_{t-1}$ for both robots. {\textit{Actions~($a_{t}\in\mathbb{R}^{10}$)}: the learned policy uses position control for joints, with positions converted to torque via a PD controller  {\small{$\tau = K_p (\hat{q} - q) + K_d (\dot{\hat{q}} - \dot{q})
$} }with fixed gains ($K_p$ and $K_d$ follow default configuration of Unitree).
{\textit{Training}}: for both oracle and unified policy training, we use PPO~\cite{DBLP:journals/corr/SchulmanWDRK17}, supported by Dagger~\cite{DBLP:journals/jmlr/RossGB11} and Actor-Critic~\cite{DBLP:conf/nips/KondaT99} for privileged learning. Rewards follow those introduced in method section, with additional basic elements from~\cite{DBLP:conf/icra/ChengSAP24, DBLP:journals/corr/abs-2404-05695}. All physics simulations perform in Isaac Gym simulator~\cite{DBLP:conf/nips/MakoviychukWGLS21}.

\noindent\textbf{Metrics.} We evaluate models based on three core criteria with levels of granularity: \locul{\textit{goal completeness}}, {\shareul{\textit{safeness}}}, and {\navul{\textit{efficiency}}}. Concretely, we use $6$ evaluation metrics -- $(1)$ {\locul{\textit{Goal Completeness}}}: Success Rate ($\%$) measuring the percentage of episodes where robots reach the hiking endpoint; Trail Completion ($\%$) indicating the portion of the trail route a robot passed; and Traverse Rate ($\%$) reflecting the distance from robot's final position (if not complete goal) to endpoint relative to total trail length. $(2)$ {\shareul{\textit{Safeness}}}: MEV ($\%$) assessing foot-edge collisions; and TTF $(seconds)$ evaluating robot stability based on episode duration before a fall occurs. $(3)$ {\navul{\textit{Efficiency}}}: Time-to-Reach~$(seconds)$ measuring average time required for successful episodes to reach endpoint. Unless specified, experiments are conducted with $512$ randomly spawned robots over $30$ seconds on $5$ distinct trail types, each featuring $5$ difficulty levels. Results are averaged over $5$ runs to minimize random biases and verify robustness.

\subsection{Ablation Study}\label{ablation}

\noindent\textbf{Settings.} We compare full LEGO-H with following designs: $(1)${\textit{Oracle}}: trained with access to privileged info and expert-designed navigation goals, representing an upper-bound performance.$(2)$ {\textit{w TC-ViT}}: LEGO-H trained without Hierarchical Latent Matching (HLM) loss metric. $(3)${\textit{Vanilla}}: LEGO-H variant where TC-ViT is replaced by a ConvGRU to predict latent and goal, altering the navigation mechanism. {\textit{We draw key observations here. Refer to {\underline{Appendix}} for more detailed comparisons and analysis.}}

\begin{table}[t!]
\small
\caption{\textbf{Ablation of LEGO-H's main components on H1.} \legendsquare{colorbestLD} for best goal completeness; \legendsquare{colorbestNE} for most safeness; \legendsquare{colorbestNC} for best efficiency.}
\vspace{-0.2in}
\label{tab:ablation_h1}
\begin{center}
\resizebox{0.47\textwidth}{!}{
    \begin{tabular}{c|c|c|c|c}
        \toprule
        \textbf{Metrics} & Oracle & LEGO-H & w TC-ViT & Vanilla  \\
        \midrule
        Success Rate (SR) (\%) $\uparrow$ &$71.20\pm0.72$ &\colorbestLD{$68.40\pm1.34$} &$64.73\pm2.22$ & $42.97\pm0.67$ \\
        Trail Completion (TC) (\%) $\uparrow$ &$77.73\pm0.92$  & \colorbestLD{$52.78\pm1.30$}&$52.50\pm1.52$ & $32.01\pm0.61$ \\
        Traverse Rate (TR) (\%) $\uparrow$ &$73.60\pm0.81$ & $71.96\pm2.37$ &\colorbestLD{$72.04\pm0.98$} &$60.26\pm0.94$  \\
        MEV (\%) $\downarrow$ &$7.12\pm0.92$  & \colorbestNE{ $7.84\pm0.92$}& $10.40\pm1.50$&  $9.41\pm1.27$\\
        TTF (s) $\uparrow$ & $7.25\pm0.09$&\colorbestNE{ $7.46\pm0.17$}& $7.00\pm0.20$& $5.36\pm0.10$ \\
        T2R (s) $\downarrow$ &$4.59\pm0.08$ &\colorbestNC{$4.95\pm0.12$} &$5.13\pm0.12$ &$6.50\pm0.07$  \\
        \bottomrule
    \end{tabular}
}
\vspace{-0.25in}
\end{center}
\end{table}

\noindent\textbf{Results.} Tab~\ref{tab:ablation_h1} indicates several insights. $(1)$ {\textit{TC-ViT is essential for basic hiking functionality.}} The consistent, significant performance advantage of {\textit{w TC-ViT}} over {\textit{Vanilla}} across all metrics, except MEV, reveals the essence of balancing the goal, physical state, and visual perception, which is crucial for coordination between navigation and locomotion.$(2)$ {\textit{Structural action behavior helps more efficient goal accomplishment and better stability.}} The absence of HLM ({\textit{w TC-ViT)} results in behaviors that complete tasks but compromise stability, often leading to mechanical risks (worse MEV than others). Including HLM ({\textit{LEGO-H)} ensures coordinated joint actions that align with the robot’s physical structure, promoting both task success (SR rises from {$64.73\%$ to $68.40\%$) and mechanical integrity (MEV goes from $10.40\%$ to $7.84\%$, TTF increase to $7.46s$), leading to more efficient task accomplishment (T2R improves from $5.13s$ to $4.95s$). $(3)${\textit{
LEGO-H rivals oracle in efficiency and safety.}} Compared to oracle which has perfect observation conditions and expert navigation goals, LEGO-H falls behind on success rate and trail completion. But surprising aspects are the efficiency and safeness, where LEGO-H's performances are comparable to or slightly better than oracle. This stresses again LEGO-H's effectiveness and capacity.

\subsection{Emerged Behaviors in Different Situations}\label{behaviors}
We further explore the behaviors that emerge in humanoid robots to unfold how robots autonomously adapt their motor skills and decision-making in response to various factors.
\begin{figure}[t!]
    \centering
    \includegraphics[width=1\linewidth]{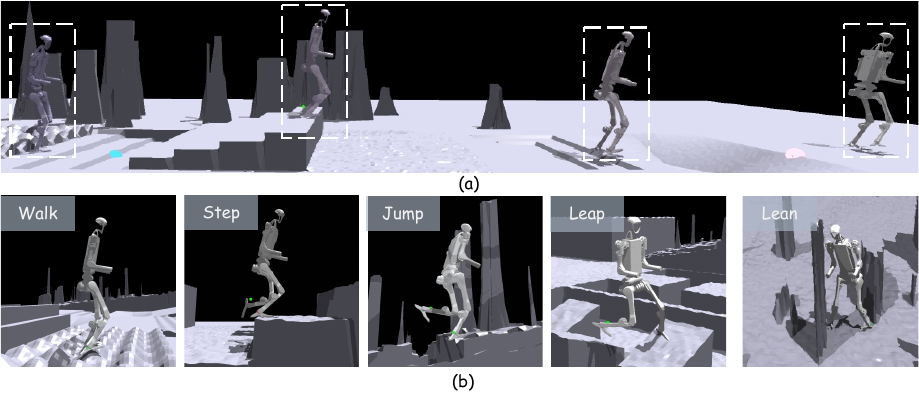}
    \vspace{-0.3in}
    \caption{\textbf{Locomotion in diverse trail terrains.} Robots developed distinct motor skills to tackle different terrains, {\textit{e.g.,}} walking on rough surfaces/leaping across ditches/leaning away high obstacles.
    }
    \vspace{-0.15in}
    \label{fig:loco-diff}
\end{figure}

\noindent\textbf{Locomotion in diverse trail terrains.} Different terrains trigger distinct locomotion behaviors, like {\textit{walking, stepping, jumping, leaping, and leaning}} (Fig~\ref{fig:loco-diff}). Key observations include: $(1)$ H1 robots typically opt for a walking gait on continuous surfaces, regardless of variations in friction, adjusting their body tilt as needed to maintain balance (Fig.~\ref{fig:loco-diff}a). $(2)$ Irregular surfaces, like fractured or sloped terrains, prompt gaits like stepping, jumping, or leaping, depending on slope and gap size (Fig.~\ref{fig:loco-diff}b). $(3)$ In tight spaces, such as cracks between large obstacles, H1's adapt by leaning sideways to navigate through these confined areas (\textit{Lean} in Fig.~\ref{fig:loco-diff}b).

\begin{figure}[t]
    \centering
    \includegraphics[width=1\linewidth]{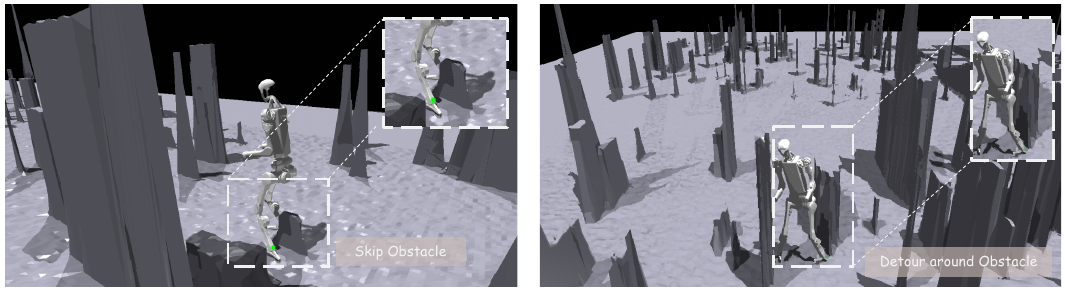}
    \vspace{-0.25in}
    \caption{\textbf{Navigation in diverse situations.} Robots developed different navigation skills, such as directly skipping a small obstacle and detouring around a high obstacle to edge through.
    }
    \vspace{-0.15in}
    \label{fig:exp-nav}
\end{figure}

\noindent\textbf{Navigation in blocked paths.} Two key behaviors are evident from Fig~\ref{fig:exp-nav}: $(1)$ When faced with tall or large obstacles, the robots typically choose to detour, maintaining a safe clearance from the obstacles. $(2)$ For obstacles below hip height, the robots initially attempt to stride or step over; if unsuccessful, they then choose to detour. These phenomena reveal the embodied character in high-level decisions. 

\begin{figure}[t]
    \centering
    \includegraphics[width=1\linewidth]{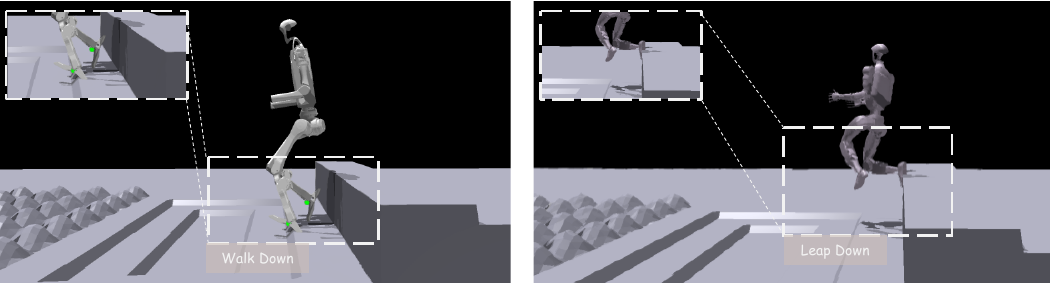}
    \vspace{-0.25in}
    \caption{\textbf{Motor behavior differences between robots.} Robots with different structures developed unique skills -- H1, which is higher and heavier, chooses to ``walk down'' step, while G1, which is shorter and more lightweight, chooses to ``leap down'' the step.
    }
    \vspace{-0.1in}
    \label{fig:exp-2}
\end{figure}

\noindent\textbf{Motor behavior differences between robots.} As shown in Fig~\ref{fig:exp-2}, when encountering identical trails like transitions between platform and flat ground, H1 and G1 exhibit different behaviors. H1 navigates down smoothly, while G1 bends its knees to jump down. This difference highlights the impact of physical mechanisms on emergent motor styles.

\subsection{Humanoid Hiking Benchmark}\label{benchmark}

\noindent\textbf{Settings.} Since current research does not directly support humanoid hiking, we selected two representative quadruped pipelines, adapting them to this task using the same input structure and oracle policy as LEGO-H. This setup allows us to investigate several key factors essential for effective humanoid hiking. The first adapted pipeline, {\textit{EP-H}}, represents a modified humanoid-hiking version of EP~\cite{DBLP:conf/icra/ChengSAP24}. The main methodological difference between EP-H and LEGO-H is that EP-H handles visual-aware navigation and locomotion by processing each depth frame independently, disregarding farther depth data to avoid distributional shifts. {\textit{RMA-H}} and {\textit{RMA-B}} are the adapted pipeline from RMA~\cite{kumar2021rma} -- the former has vision inputs, and the later is blind.  This pipeline originally supports blind locomotion, and employs a frozen oracle policy with an adapter network to map real-world sensory data to oracle’s latent space for policy adaptation. 

\begin{table}[t!]
\small
\caption{\textbf{Humanoid hiking benchmark for H1 across all trail categories.}  \legendsquare{colorbestLD}/\legendsquare{colorbestNE}/\legendsquare{colorbestNC} show best goal completeness/safeness/efficiency. }
\vspace{-0.2in}
\label{tab:hiking_benchmark}
\begin{center}
\resizebox{0.485\textwidth}{!}{
    \begin{tabular}{c|c|c|c|c}
        \toprule
        \textbf{Metrics} & \textbf{LEGO-H} & \textbf{EP-H} & \textbf{RMA-H} & \textbf{RMA-B} \\
        \midrule
        Success Rate (\%) $\uparrow$ & \colorbestLD{$68.40\pm1.34$} & $28.80\pm0.88$ & $65.17\pm2.05$ & $48.11\pm0.72$ \\
        Trail Completion (\%) $\uparrow$ & \colorbestLD{$52.78\pm1.30$} & $25.98\pm0.22$ & $52.51\pm1.41$ & $41.92\pm0.34$ \\
        Traverse Rate (\%) $\uparrow$ & $71.96\pm2.37$ & $64.16\pm0.48$ & \colorbestLD{$74.61\pm0.93$} & $69.85\pm1.50$ \\
        MEV (\%) $\downarrow$ & \colorbestNE{$7.84\pm0.92$} & $12.44\pm1.32$ & $8.70\pm1.55$ & $10.74\pm1.13$ \\
        TTF (s) $\uparrow$ & \colorbestNE{$7.46\pm0.17$} & $4.64\pm0.13$ & $6.97\pm0.17$ & $5.22\pm0.03$ \\
        Time-to-Reach (s) $\downarrow$ & \colorbestNC{$4.95\pm0.12$} & $9.79\pm0.16$ & $4.98\pm0.11$ & $6.19\pm0.05$ \\
        \bottomrule
    \end{tabular}
}
\vspace{-0.15in}
\end{center}
\end{table}

\noindent\textbf{Results.} We focus on three vital questions from the benchmark: $1)$ \textit{Is {\underline{visual perception} essential for integrated navigation and locomotion?}} $2)$ \textit{What type of {\underline{visual information} is most effective?}} $3)$ \textit{Is {\underline{unified}} cross-level learning necessary?} Key findings in Tab~\ref{tab:hiking_benchmark} and visualizations in Appendix revel the answers:$(1)$ {\textit{Vision is essential.}} Without vision, RMA-B struggles across all metrics, highlighting the need for visual feedback. $(2)$ {\textit{Goal-aligned, multi-scale visual perception is critical.}} \textit{EP-H}, which processes each depth frame independently without continuous goal alignment, and brute-force cutoff distance information, results in frequent circles and fails to lock onto navigation paths. The performance gap between LEGO-H and EP-H across metrics underscores the importance of structured visual information. $(3)$ {\textit{Unified learning is vital for adaptability.}} \textit{RMA-H} performs adequately on straight paths but fails with turns or obstacles, showing that locomotion feedback alone is insufficient for embodied-aware decision-making. A unified learning framework supports essential cross-level interaction, enabling adaption and effectiveness across all levels.

\section{Conclusion}

We propose {\textit{humanoid hiking}} as a new testbed for advancing research in embodied autonomy. To address the challenges it poses, we introduce LEGO-H, a unified policy learning framework that highlights the importance of integrative skill development for a humanoid to autonomously accomplish complex tasks like hiking. Experiments demonstrate effectiveness of LEGO-H and also uncover promising directions for future research, like whole-body control, long-horizon exploration, and visual-motor coordination. 

\subsection*{Acknowledgments}
This project was supported, in part, by NSF 2215542, NSF 2313151, and UM U081626 to S. Yu.  

\clearpage
{
    \small
    \bibliographystyle{ieeenat_fullname}
    \bibliography{main}
}
\clearpage

\appendix

\title{
\noindent\Large\textbf{{Appendix}}}

\maketitle

\begin{abstract}
In the appendix, we provide a comprehensive elaboration of LEGO-H. Section~\ref{position} recaps the positioning of the Humanoid Hiking task and highlights how LEGO-H departs from the current trends in humanoid robotics. Section~\ref{supp_ref} expands on related work. Section~\ref{additional_ablation_h1} delves into extended ablation studies, analyzing detailed design choices of each component in LEGO-H. Section~\ref{uni} explores the framework's universality through experiments on the integration of LEGO-H components into alternative frameworks. Section~\ref{sim_env} introduces the simulated environments developed for training and evaluation in this new hiking paradigm. Section~\ref{supp_exp} specifies implementation details. Section~\ref{supp-benchmark} extends evaluations on critical questions in humanoid hiking. Lastly, section~\ref{future_work} discusses future work. 

\end{abstract}

\section{The Positioning of LEGO-H}\label{position}

To better understand LEGO-H’s positioning, we present a conceptual framework comparison in Fig~\ref{position}. LEGO-H advances humanoid robotics by seamlessly integrating navigation and locomotion into a unified policy learning framework (Fig.~\ref{position}c). This contrasts with existing pipelines, which either separate these modules (Fig.~\ref{position}a) or reduce environmental complexity by relying on external commands for action execution (Fig.~\ref{position}b).

This work emphasizes the importance of integrative development of navigation and locomotion for humanoid robots to operate effectively in complex real-world environments. Humanoid hiking provides an ideal testbed to evaluate this coordination. LEGO-H, as a baseline prototype, demonstrates how unified learning fosters emergent behaviors, enabling dynamic adaptation to diverse trails and challenges.
\begin{figure}[t!]
    \centering
    \includegraphics[width=1\linewidth]{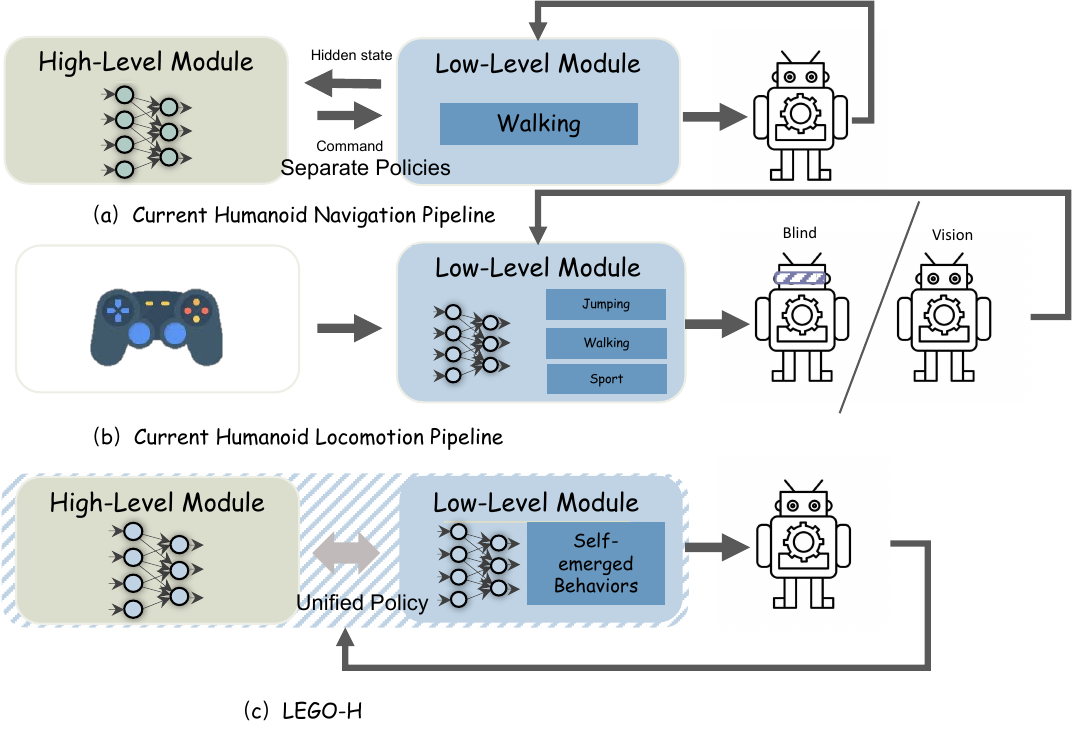}
    \vspace{-0.2in}
    \caption{\textbf{The conceptual framework differences.} We summarize the key conceptual level differences between our work and current humanoid robot trends for better positioning of LEGO-H. }
    \label{fig:position}
\end{figure}

\section{Additional Related Work }\label{supp_ref}
\subsection{Hierarchical RL}
 It is widely adopted to decompose a complex RL problem into multiple layers of policies~\cite{DBLP:conf/nips/DayanH92,DBLP:journals/ai/SuttonPS99}. This paradigm naturally structures in hierarchy, where a decision-making/control module at higher levels manages temporal (longer time scale) and behavioral abstraction, while a low-level module focuses on atomic skills to execute momentary actions in the environment, guided by the high-level module. HRL includes two main methodologies: $(1)$ explicit goal setting~\cite{DBLP:conf/nips/NachumGLL18}, where the high-level policy assigns target goals to the low level, enhancing reusability but limiting adaptability, and $(2)$ latent space policies~\cite{DBLP:journals/scirobotics/LeeBRWMH24}, where high-level module guides the low-level policy by providing latent sub-goals at a lower frequency, offering flexibility but often limiting generalization. However, HRL are generally not end-to-end trainable due to complexity and distinct objectives of each level. Our {\textit{LEGO-H}}, is also hierarchical but avoids strict goal adherence or explicit skill definitions. Instead, it presents a unified, end-to-end policy learning framework, where high-level module offers latent representations and intermediate goals as flexible guidance, allowing low level to reference them adaptively rather than following rigidly. This soft guidance supports adaptability and coherence in complex environments, addressing traditional HRL limitations.

 \subsection{Privileged Learning}
 It is a two-stage technique in robotics, often employed to address sim-to-real transfer challenges~\cite{DBLP:journals/jmlr/VapnikI15a, DBLP:journals/scirobotics/HwangboWK020,DBLP:conf/corl/0001ZKK19}. For first {\textit{teacher}} stage, the robot agent learns an {\textit{oracle} policy via additionally accessing privileged information from human demonstrations~\cite{DBLP:conf/corl/0001ZKK19}, or GT exteroceptive measurements from simulator~\cite{DBLP:journals/scirobotics/HwangboWK020}. Since extra information reduces ambiguity via precise physical states/terrain details/expert trajectories, the agent could learn more precise actions. However, as this information is unavailable in real-world deployment, in the second {\textit{student}} stage, the robot agent learns to imitate the teacher’s behavior using only accessible data~\footnote{It often includes proprioception, user commands, and visual sensor inputs.} through knowledge distillation. Common distillation losses target element-wise difference~\cite{DBLP:conf/corl/0001ZKK19}, distribution alignment~\cite{2022-TOG-ASE} or latent space alignment~\cite{kumar2021rma}. However, studies rarely address the structural consistency of actions, a critical factor for humanoid hiking, where the robot’s high articulation requires precise coordination across joints. 

\section{Additional Ablation Studies}
\label{additional_ablation_h1}

In this section, we delve into detailed designs of TC-ViT (Sec.~\ref{TC-VITs-detials}) and Hierarchical Latent Matching (HLM) loss metric (Sec.~\ref{HLM-detials}). Additionally, we example and analyze further emergent behaviors focusing on the {\textit{safeness}} aspect (Sec.~\ref{eba}), which were not covered in the main paper.
\subsection{Efficiency of TC-ViT}\label{TC-VITs-detials}
In this subsection, we further analyze the efficiency behind TC-ViTs' recurrent goal adaptation module design.

\noindent\textbf{Why Recurrent Goal Adaptation?} As mentioned in the main paper, this module, implemented via a GRU and grafted at the end of TC-ViT -- integrates motor actuation and physical body states, enhancing visual cue processing with proprioceptive insight. While recent advances like CausalTransformers (CTs)~\cite{DBLP:conf/corl/Zeng0EHSHGKW24,DBLP:journals/corr/abs-2303-03381} have shown promising results in temporal modeling, we intentionally adopt a GRU-based design due to its better computational efficiency: TC-ViT has Flops-$0.686G$/Params-$31.25M$, while replacing its GRU to CTs increase to $0.785G/55.92M$. Besides, CTs require significantly more computational resources for sufficient training, leading to performance degradation under the same memory constraints (Tab.~\ref{tab:ablation_gru}). Since most visual information is already processed by the preceding ViViT-style encoder, CTs would introduce redundancy in such a later stage. An additional finding is that our HLM helps improve CTs performance—{\textit{e.g.}}, reducing CT’s collision (MEV) from $10.48\%$ to $8.61\%$.

\begin{table}[h!]
\caption{\textbf{GRU vs CTs at the end of TC-ViT.} }
\label{tab:ablation_gru}
\vspace{-1.5ex}
\resizebox{0.47\textwidth}{!}{
    \begin{tabular}{c|c|c}
        \toprule
        \textbf{Metrics} & w GRU &  w CTs \\
        \midrule
        Success Rate (\%) $\uparrow$ & \colorbestLD{$68.40\pm1.34$} &  $27.85\pm1.02$\\
        TTF (s) $\uparrow$ & \colorbestNE{$7.46\pm0.17$} & $5.44\pm0.34$ \\
        \bottomrule
    \end{tabular}}
\end{table}

\subsection{How HLM Works}\label{HLM-detials}

In this subsection, we further analyze the Hierarchical Latent Matching (HLM) loss metric by addressing two key questions: $(1)$ {\textit{How does the structural rationality of actions impact the safety of the robot's movements?}} $(2)$ {\textit{Is a vanilla VAE sufficient to capture and reflect the rationality of the robot's actions?}} Through these investigations, we aim to provide deeper insights into the design choices and contributions of HLM for promoting self-coordinated and safe humanoid movements across complex trails.

\noindent\textbf{Ablation on w/wo HLM.} We show the quantitative comparison between w/wo HLM in Tab {\color{red}1} of the main paper with metric {MEV}. Here, as a complementary, we show qualitative samples. As shown in Fig~\ref{fig:ablation-hlm}, while LEGO-H without HLM achieves successful traversal over the hurdle, the mechanical risks are significantly higher. The robot's right leg collides with the hurdle during the stepping motion, and the minimal clearance further demonstrates unsafe and inefficient movement patterns. In contrast, with HLM incorporated, the robot executes structurally rational and safe movements. It first steps onto the hurdle with its left leg, ensuring sufficient clearance for the right leg, and then transitions to a stable hop onto the opposite leg. This coordinated behavior highlights the role of HLM in enabling stability, safety, and effective traversal strategies. 

\begin{figure}[t!]
    \centering
    \includegraphics[width=1\linewidth]{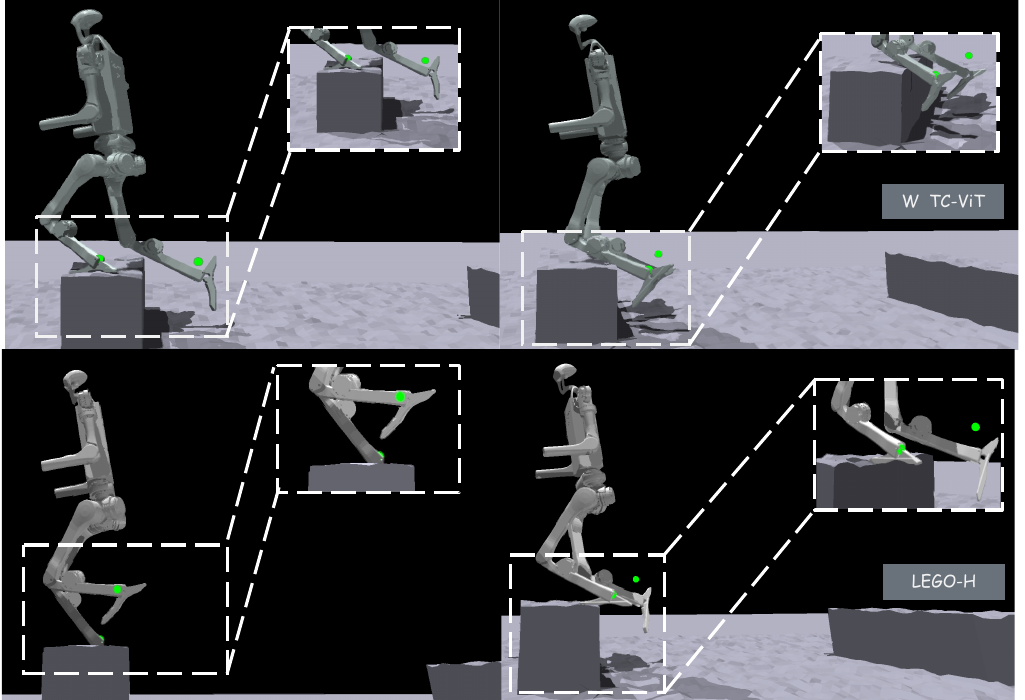}
    \vspace{-0.2in}
    \caption{\textbf{Qualitative ablation on with/without HLM.}  Snapshots from right to left depict two time steps of a robot traversing a hurdle obstacle. The top row illustrates behaviors without HLM, where unsafe movements lead to right leg collisions with the hurdle. The bottom row showcases behaviors with HLM, exhibiting coordinated and structurally rational actions that ensure stability and successful traversal with safe clearance.}

    \label{fig:ablation-hlm}
\end{figure}

\noindent\textbf{Vanilla VAE or full HLM?}  The latent space of a vanilla VAE is commonly employed for prior regularization, promoting outputs that align with the normal distribution of the data. This proves effective for tasks like approximating averages in large-scale or in-the-wild datasets, as seen in human pose reconstruction~\cite{SMPL-X:2019}. However, vanilla VAE falls short when structural dependencies and inter-joint dynamics are critical, like humanoid robot actions. Specifically, humanoid hiking with safety demands fine-grained understanding of hierarchical relationships of robots' own physical mechanism, which vanilla VAE lacks. By contrast, as demonstrated in Tab~\ref{tab:vae}, full HLM introduces additional masked reconstruction and hierarchical losses that implicitly enforce inter-joint structural rationality, enabling safer and more efficient robot movement in complex tasks like humanoid hiking.
\begin{table}[h!]
\small
\caption{\textbf{Ablation of HLM.}  \legendsquare{colorbestLD} for best goal completeness; \legendsquare{colorbestNE} for most safeness; \legendsquare{colorbestNC} for best efficiency. The results highlight the insufficiency of using a vanilla VAE as a prior. Additionally, compared with Tab.~{\color{red}1} in the main paper, the vanilla VAE collapses actions into average motions. While this slightly improves MEV compared to the setting without any prior (w TC-ViT), it sacrifices performance across all other metrics.}
\vspace{-0.1in}
\label{tab:vae}
\begin{center}
\resizebox{0.47\textwidth}{!}{
    \begin{tabular}{c|c|c}
        \toprule
        \textbf{Metrics} & full HLM & Vanilla VAE \\
        \midrule
        Success Rate (\%) $\uparrow$ &\colorbestLD{$68.40\pm1.34$} &$53.49\pm1.61$ \\
        Trail Completion (\%) $\uparrow$ & \colorbestLD{$52.78\pm1.30$}&$43.00\pm0.96$  \\
        Traverse Rate (\%) $\uparrow$ & \colorbestLD{$71.96\pm2.37$}& $64.52\pm1.02$ \\
        MEV (\%) $\downarrow$ & \colorbestNE{$7.84\pm0.92$}& $9.26\pm1.08$  \\
        TTF (s) $\uparrow$ &\colorbestNE{$7.46\pm0.17$} &$6.30\pm0.15$   \\
        Time-to-Reach (s) $\downarrow$ &\colorbestNC{$4.95\pm0.12$} & $6.02\pm0.05$ \\
        \bottomrule
    \end{tabular}
}

\end{center}
\end{table}
\subsection{Emergent Behavior Analysis}\label{eba}
In this subsection, we explore a critical question: \textit{How do robots behave to ensure safety?} We will list three examples, considering both high-level navigation behaviors and low-level motor skill execution, to show how LEGO-H prioritizes safety in dynamic and challenging environments.
\begin{figure}[t!]
    \centering
    \includegraphics[width=1\linewidth]{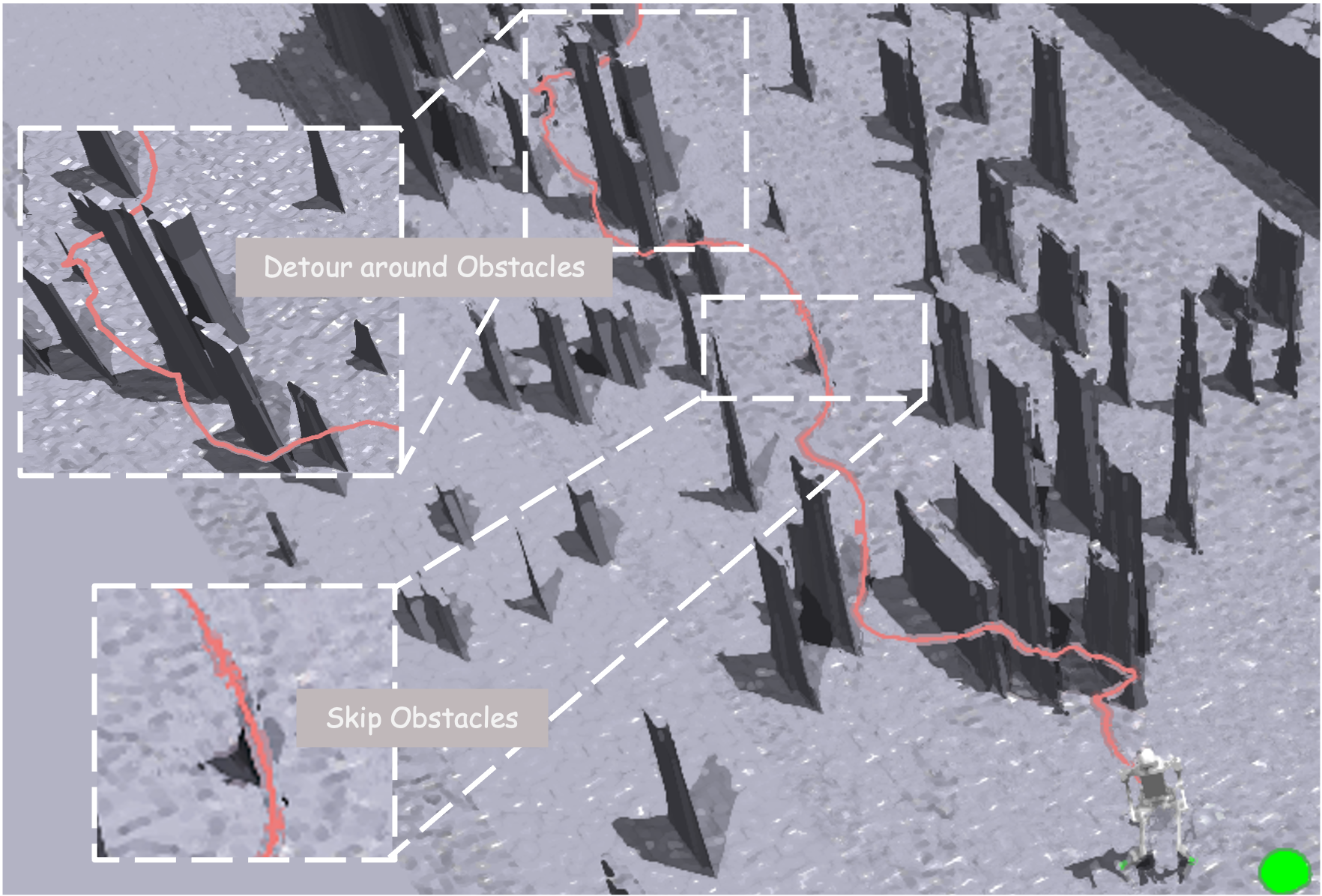}

    \caption{\textbf{Navigation in blocked paths over different obstacles.} The colored trajectory illustrates the robot's torso position as it traverses the trail. Zoomed-in regions highlight distinct navigation behaviors: when encountering crowded, tall obstacles, the robot opts to detour, whereas for smaller obstacles, the robot leaps over, demonstrating adaptive navigation strategies.}
\vspace{-0.2in}
    \label{fig:nav-path-behave}
\end{figure}

\noindent\textbf{Navigation in blocked paths.} As discussed in the main paper Section~{\color{red}4.3}, robots typically opt to detour around large, tall obstacles and skip over smaller ones. Here, we show the phenomena from another aspect. In Fig~\ref{fig:nav-path-behave}, the traversed trajectory shows substantial clearance maintained from tall obstacles (zoomed-in block: detour over obstacles) and efficient traversal above smaller ones (zoomed-in block: skip obstacles). This demonstrates the robot's ability to prioritize collision avoidance while exhibiting adaptive decision-making based on the encountered environment.

\noindent\textbf{Behavior over different terrains.} In the main paper Section~{\color{red}{4.3}}, we discussed how diverse and distinct locomotion skills emerge to tackle different terrains. Here, we present two examples demonstrating how terrains influence the robots' integrative navigation decisions and motor execution. As shown in Fig~\ref{fig:diff-walk}: $(1)$ for a {\textit{long, rugged}} gully, the robot adopts a lateral "crab walk" strategy to maintain balance and progress towards the terminus. (2) For a {\textit{short, smooth}} gully, the robot directly faces the terminus and leaps over it, showcasing adaptive integrative navigation and motor behavior responses to varying trail challenges.

\begin{figure}[t!]
    \centering
    \includegraphics[width=1\linewidth]{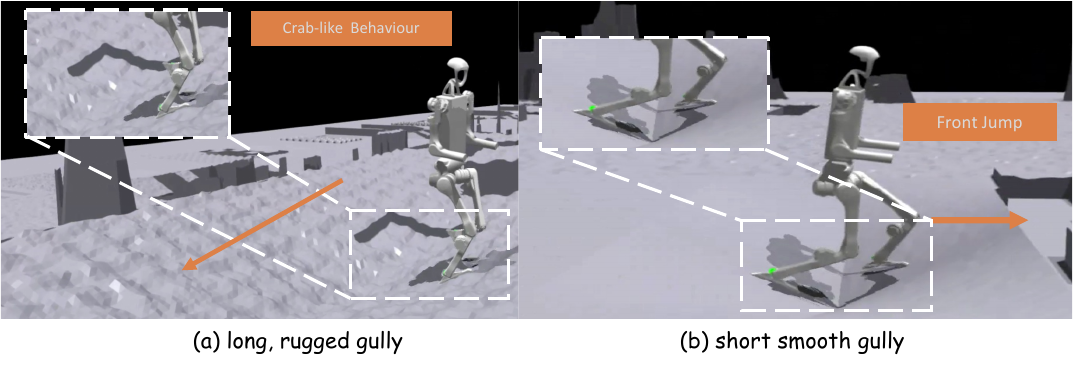}
    \vspace{-0.2in}
    \caption{\textbf{Behaviors over difference terrains.} The robots exhibit diverse integrative navigation and locomotion skills tailored to varying trail terrains. (a) The robot adopts a lateral "crab" walking style to navigate a long, rugged gully, maintaining stability while progressing toward the hiking terminus. (b) The robot faces the final terminus directly and jumps over a short, smooth gully. The orange directional lines show the terminus directions. }
    \label{fig:diff-walk}
      \vspace{-0.1in}
\end{figure}

\begin{figure*}[t!]
    \centering
    \includegraphics[width=1\linewidth]{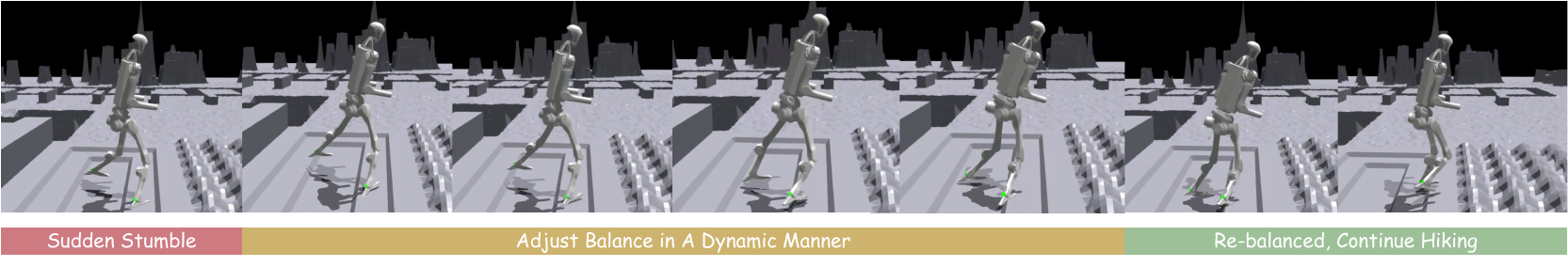}
\vspace{-0.2cm}
    \caption{\textbf{Self re-balance.} The robot stumbles unexpectedly (red timeline), swiftly adjusts its balance through a sequence of emergent lateral motions ({yellow timeline), and seamlessly regains stability (green timeline). }}

    \label{fig:reblance}
\end{figure*}
\noindent\textbf{Re-balancing.} The ability to re-balance is critical for humanoid robots traversing complex trails. As shown in Fig~\ref{fig:reblance}, the robot stumbles due to uneven terrain (red timeline), triggering a sequence of emergent lateral motions that dynamically counteract the imbalance (yellow timeline). After that, the robot shows seamless coordination between re-balancing and task continuity (green timeline). This example highlights that, rather than relying on predefined recovery motions, the robot adapts its behavior dynamically to the context. Such adaptability underscores the robustness of LEGO-H’s unified learning framework in fostering emergent, and context-aware integrative navigation and motor skills with safeness.

\section{The Universality of LEGO-H}\label{uni}
 In this section, we explore the universality of LEGO-H by demonstrating its flexibility in two ways: $(1)$ integrating key components like HLM into other policy learning pipelines, and $(2)$ transferring the entire framework to a morphologically distinct humanoid robot, Unitree G1, without architecture changes.
\subsection{HLM as a Plug-in Supervision }
HLM focuses exclusively on maintaining structural similarity between the oracle locomotion policy's actions and the student's, making it agnostic to the student's framework design. This modularity allows HLM to be seamlessly integrated as a plug-in supervision component into different policy architectures, ensuring structural rationality and coordination without requiring changes to the underlying framework. We demonstrate this property by adding it to  {EP-H}. The results are shown in Tab~\ref{plug}.
\begin{table}[h!]
\small

\caption{\textbf{HLM as a plug-in supervision for other framework.} }\label{plug}

\vspace{-0.12in}
\label{tab:ablation_h1}
\resizebox{0.47\textwidth}{!}{
    \begin{tabular}{c|c|c}
        \toprule
        \textbf{Metrics} &   EP-H & EP-H + HLM \\
        \midrule
        Success Rate (SR) (\%) $\uparrow$   &$28.80\pm0.88$  & $35.53\pm1.30$ \\
        Trail Completion (TC) (\%) $\uparrow$   & $25.98\pm0.22$ &  $30.36\pm0.89$ \\
        Traverse Rate (TR) (\%) $\uparrow$  &$64.16\pm0.48$  & $58.23\pm0.76$  \\
        MEV (\%) $\downarrow$   & $12.44\pm1.32$ & $10.98\pm1.40$ \\
        TTF (s) $\uparrow$  & $4.64\pm0.13$ & $5.04\pm0.16$ \\
        T2R (s) $\downarrow$   & $9.79\pm0.16$ & $7.80\pm0.37$ \\
        \bottomrule
    \end{tabular}
}
\end{table}

\subsection{Transfer to G1 Robot }
To further evaluate the universality of LEGO-H, we retrain the framework on the Unitree G1 humanoid robot without any architectural modification — demonstrating its agnosticism to specific robot morphology. As shown in Tab~\ref{tab:g1}, two key observations emerge from this transfer: $(1)$ Framework generalization: LEGO-H can adapt to G1, despite differences in body structure and joint configuration from H1. LEGO-H on G1 preserves reasonable integrative navigation and locomotion performance. $(2)$ Performance shift. Compared to H1, G1 exhibits lower performance in general. This is primarily due to its shorter leg length and reduced camera height, which constrain both physical reach and perceptual field. Thus, on tasks requiring large clearance—such as jumping over ditches, G1 typically struggles more. A possible solution to mitigate these limitations is to extend LEGO-H with effective whole-body control (WBC) designs, allowing more expressive coordination across the upper body and the lower body. This could compensate for morphological constraints and unlock more agile, full-body responses to complex hiking trails.
\begin{table}[h!]
\small
\caption{\textbf{LEGO-H on Humanoid G1 robot.} We list H1's result as a reference. The results highlight the universality of our proposed learning framework for different robot types.}
\vspace{-0.1in}
\label{tab:g1}
\resizebox{0.47\textwidth}{!}{
    \begin{tabular}{c|c|c}
        \toprule
        \textbf{Metrics} & H1 & G1 \\
        \midrule
        Success Rate (\%) $\uparrow$ &$68.40\pm1.34$ &$63.96\pm1.03$ \\
        Trail Completion (\%) $\uparrow$ & $52.78\pm1.30$& $38.94\pm0.63$ \\
        Traverse Rate (\%) $\uparrow$ & $71.96\pm2.37$& $62.21\pm0.97$ \\
        MEV (\%) $\downarrow$ & $7.84\pm0.92$& $5.33\pm0.68$  \\
        TTF (s) $\uparrow$ &$7.46\pm0.17$ &$7.24\pm0.22$   \\
        Time-to-Reach (s) $\downarrow$ &$4.95\pm0.12$ & $8.10\pm0.08$ \\
        \bottomrule
    \end{tabular}
}

\end{table}

\section{Simulated Hiking Trail Constructions}\label{sim_env}
To establish a robust testbed for humanoid hiking tasks, we design diverse trails in the Nvidia Isaac Gym Simulator~\cite{DBLP:conf/nips/MakoviychukWGLS21} using a procedural generation approach. The construction process is detailed in Section~\ref{trail-gen}, while Section~\ref{goal-design} outlines the goal and waypoint design methodology. 
\subsection{Trail Scene Generation}\label{trail-gen}
To simulate diverse trail environments for humanoid hiking, we design $16$ basic terrain primitives. Each primitive is extended into multiple variants by randomly sampling terrain properties such as slope, height, and surface friction, as well as their positions, using a procedural terrain generation mechanism. These primitives form the foundation for constructing five distinct trail types, each presenting a unique combination of terrain challenges and navigation complexity. Specifically:
\begin{itemize}
    \item {\textit{RandomMix}} trail category features unobstructed views, testing the robot's ability to navigate long distances while adapting multiple motor skills to various mixed terrain types.
    \item {\textit{Ditch}} category introduces uneven, middle-distance trails with diverse slopes and gaps, challenging the robots to decide and execute quick turns and agile leaps.
    \item {\textit{Hurdle}} category includes trails with long, cubic obstacles, focusing on testing the robot's ability to avoid foot collisions while navigating middle distances. 
    \item {\textit{Gap}} trails with uneven jumping platforms, including varying gap distances and straight or staggered stones, evaluating the robot's balance and jumping ability during middle-distance navigation.
    \item {\textit{Forest}} trails densely populated with variously sized and positioned obstacles, simulating obstructed views and tight navigation spaces. These test the robot's ability to detour, effectively traverse crowded paths, and maintain balance under constrained conditions.
\end{itemize}

Each trail category covers five hiking difficulty levels, with additional variants generated through the randomization of terrain properties and obstacle placement. These diversities ensure a comprehensive testbed across a wide spectrum of challenges. To expand the evaluation scope, we also construct out-of-domain hiking trails by combining multiple trail types into complex, long-distance hill scenarios. These trails test the robots’ adaptability, and integrative capabilities under extended and unpredictable hiking conditions. We show the zero-shot ability of LEGO-H on the out-of-domain trails in the supplemental video.

\subsection{Oracle Navigation Goal Design}\label{goal-design}
The design of expert navigation goals for the oracle stage follows these criteria:

\begin{itemize} 
\item \textit{Unobstructed-view trails}: For trails with clear visibility, such as \textit{RandomMix}, expert navigation goals are set as evenly spaced waypoints within the traversable regions, aligning directly with the trail direction. These goals ensure smooth long-distance navigation. 
\item \textit{Obstructed-view trails}: For complex trails like \textit{Forest}, navigation goals are dynamically set to detour around obstacles, following feasible paths with a degree of randomness to promote diverse path exploration. These goals maintain sufficient clearance to prevent collisions and encourage obstacle-aware navigation strategies. 
\item \textit{Terrain-specific trails}: For specialized challenges like \textit{Hurdle}, \textit{Ditch}, and \textit{Gap}, navigation goals are positioned to encourage the emergence of specific motor behaviors, such as agile leaps, balanced stepping, or jumping within safe zones. These goals are carefully tailored to meet the unique demands of each terrain type, ensuring both adaptability and safety.
\end{itemize} 
These navigation goals establish a robust foundation for oracle policy training.

\section{Experimental Details}\label{supp_exp}
All experiments are conducted on a single A40 GPU, though the policy can also be deployed on a more cost-effective GPU, such as the 4080. The oracle policy training requires approximately $\sim18$ GPU hours, while the unified policy training takes $\sim2$ GPU days. For camera placement, if the humanoid robots are equipped with a head-mounted camera, we use the default configuration. Otherwise, an additional camera is attached approximately at eye level. This section provides additional implementation details of LEGO-H: Section~\ref{arc} details the architecture specifications, and Section~\ref{tp} elaborates on the training procedures and hyperparameter configurations.
\subsection{Network Architectures}\label{arc}
This section details the network architectures of: the scandot encoder, the oracle policy, and the masked Variational Autoencoder (VAE) used in the Hierarchical Loss Metric (HLM).

\noindent\textbf{Scandot Encoder.} It is three layers of MLPs, with the hidden layer dimension of $[128,64,32]$. The activation functions are eLU for hidden layers and Tanh for the output layer.

\noindent\textbf{Oracle Policy.} The Actor network takes proprioceptive data, encoded scan features from the Scandot Encoder, privileged information, and encoded privileged features as inputs, and flows them into three layers of MLPs, where the dimension is $[512,256,128]$. The activation functions are eLU for hidden layers and Tanh for the output layer.The Critic network shares the same architecture as the Actor network. The encoder dimension for privileged information is $[64,20]$. 

\noindent\textbf{Masked VAE for HLM.} The architecture of the Variational Autoencoder (VAE) employed for the Hierarchical Loss Metric (HLM) consists of fully connected residual layers. The encoder includes multiple ResidualFC layers followed by two linear layers to produce the mean and log variance of the latent variable. ReLU activations are used in both the encoder and decoder, with the decoder's output layer utilizing a sigmoid activation function to ensure bounded outputs.

\subsection{Training Procedure}\label{tp}
The training process begins with the development of {\textit{oracle}} policy using privileged information and expert navigation goals. Subsequently, the unified policy, incorporating TC-ViTs and the locomotion module, is trained with visual information as inputs. This stage excludes privileged information and distills motor knowledge from the oracle policy into the unified framework.
\subsubsection{Oracle Policy Training}
The goal of this stage is to develop an oracle locomotion policy that facilitates the training of the unified policy in the subsequent stage. Since the environment properties will be unknown in the second stage, we adopt the strategy from~\cite{DBLP:conf/icra/ChengSAP24,wu2024helpful} to train an adaptation module capable of estimating environment properties. The detailed training procedure is outlined below. 

\noindent\textbf{Curriculum Learning.} To ensure stable training, we leverage curriculum learning~\cite{DBLP:journals/scirobotics/HwangboWK020,DBLP:conf/icra/ChengSAP24,DBLP:journals/scirobotics/LeeBRWMH24}, progressively increasing the complexity of traversable terrains based on the robots' acquired skills. This method enables gradual adaptation and robust policy development for challenging trails. Specifically, the robot's distance from the origin is tracked and compared against a threshold determined by its commanded velocity and the episode length. Terrain levels are adjusted as follows: $(1)$ if the robot's distance exceeds $80\%$ of the threshold, the terrain level advances to a more challenging stage; $(2)$ if the robot's distance falls below $40\%$ of the threshold, the terrain level reverts to an easier stage; and $(3)$ upon completing all levels, the robot is randomly reassigned to a level to maintain diversity in training.
\begin{table}[h!]
\caption{\textbf{Domain randomization parameters.}}\label{dr}
\centering
\begin{tabular}{c|c}
\hline
\textbf{Term} & \textbf{Value} \\
\hline
Friction & $\mathcal{U}(0.6, 2.0)$ \\
Base Mass offset & $\mathcal{U}(0.0, 3.0)$ \\
Base CoM offset &$\mathcal{U} (-0.2, 0.2)$ \\
Push robot--interval & 8s \\
Push robot--max push vel\_xy & 0.5m/s \\
Motor strength range & $\mathcal{U}(0.8, 1.2)$ \\
Delay update global steps & $24 \times8000$ \\

\hline
\end{tabular}

\end{table}
\begin{table}[h!]
\caption{\textbf{Loss weight hyperparameters.}}
\centering
\resizebox{0.46\textwidth}{!}{
\begin{tabular}{c|cccccccccc}
\hline
\textbf{Parameter} & $w_1$ & $w_2$ & $w_3$ & $w_4$ & $w_5$ & $w_6$ & $w_7$ & $w_8$ & $c_{mt}$ & $c_{ms}$ \\
\hline
\textbf{Value} & $1.0$ & $1.0$ & $1.0$ & $1.0$ & $1.0$ & $1.0$ & $100.0$ & $2.0$ & $0.85$ & $0.15$ \\
\hline
\end{tabular}
\vspace{-0.2cm}
}
\label{loss_weight}
\end{table}

\begin{table*}[h!]
\caption{\textbf{Rewards' definition and weight}.The symbol $*$ means the term only used in unified policy training stage.}
\centering
\begin{tabular}{c|c|c}
\hline
\textbf{Term} & \textbf{Mathematical Expression} & \textbf{Weight} \\
\hline
Tracking Goal Velocity & $\frac{\min\left(\mathbf{v}_{\text{target}} \cdot \mathbf{v}_{t}, \text{cmd}_x\right)}{\text{cmd}_x + \epsilon}
$
 & 10.0 \\
Tracking Yaw & $\exp\left(-\left|\psi_{\text{target}} - \psi_{t}\right|\right)
$ & 0.5 \\
Linear Velocity (Z) &$ v_z^2$ & -2.0 \\
Angular Velocity (XY) &$\sum \left(\omega_x^2 + \omega_y^2\right)
$ & -1.0 \\
Orientation &$\sum \left(g_x^2 + g_y^2\right) $ & -1.0 \\
DOF Acceleration &$ \sum \left(\frac{\dot{q}_{t-1} - \dot{q}_{t}}{\Delta t}\right)^2
$ & -3.5e-8 \\
Collision & $\sum \left(\|\mathbf{F}_{\text{contact}}\| > 0.1\right)
$& -10.0 \\
Action Rate &$\|\mathbf{a}_{t-1} - \mathbf{a}_{\text{t}}\|
$ & -0.01 \\
Delta Torques &$ \sum \left(\tau_{t} - \tau_{t-1}\right)^2
$ & -1.0e-7 \\
Torques & $\sum \tau_{t}^2
$& -1.0e-5 \\
Hip Position &$\sum \left(q_{\text{hip}} - q_{\text{hip-default}}\right)^2
$ & -0.5 \\
DOF Error &$r_{\text{dof\_error}} = \sum \left(q_{\text{dof}} - q_{\text{default}}\right)^2
$ & -0.04 \\
Feet Stumble & $\bigvee_{\left(\|\mathbf{F}_{\text{contact}}\| > 4 \cdot \left|F_{\text{contact}}\right|\right)}
$& -1 \\
Feet Edge & $ (\text{terrain\_level} > 3) \cdot \sum \left(\text{feet\_at\_edge}\right)
$& -1 \\
Feet Air Time & $\sum \left(T_{\text{air}} - 0.5\right) \cdot (\text{first\_contact})

$ & 1.0(H1)/0.5(G1) \\
Base Height &$\left(h_{\text{base}} - h_{\text{target}}\right)^2$ & -100.0 (H1)/-35.0 (G1)\\
Point Navigation Distance$^*$ &$r_{\text{pn\_distance}} = 
\begin{cases} 
    1 & \|\mathbf{p}_{\text{rel}}\| < \theta_{reach} \\
    -\|\mathbf{p}_{\text{rel}}\| \cdot 0.75 & \text{otherwise}
\end{cases}
$ & 1.0 \\

DOF Position Limits &$

\sum\left( -\max\left( 0, \text{dof} - \text{dof\_lim}_{low} \right) + \max\left( 0, \text{dof} - \text{dof\_lim}_{ up} \right) \right)$ & 0.0 (H1)/-5.0 (G1) \\

Tracking Sigma & $\exp(-track_{err}^2/\sigma)$& 0.5 \\

\hline
\end{tabular}
\label{rewards} 
\end{table*}

\noindent\textbf{Domain Randomization.} To increase the sim-to-real transfer ability, we follow the common strategy in robotics to use the ~\cite{DBLP:conf/iros/TobinFRSZA17}. The detailed parameters are listed in Tab~
\ref{dr}.

\noindent\textbf{Rewards.} Please refer to Tab~\ref{rewards} for the detailed formula definitions and corresponding weights.

\noindent\textbf{Termination Conditions.} To maintain meaningful training and testing environments, we define termination conditions to prevent invalid episodes. An episode ends if any of the following occur: $(1)$ \textit{Soft pose check}: the robot’s absolute roll or pitch exceeds a predefined threshold, or its height falls below a defined lower bound; $(2)$ \textit{Goal reach check}: the robot is within a specific distance from the final goal. We adopt the goal navigation criteria from~\cite{DBLP:journals/corr/abs-1807-06757}, setting the goal distance to roughly twice the robot’s body width. Specifically, the goal distance is set to $0.89$ during testing and $0.5$ during training to encourage precise task execution. $(3)$ {\textit{Timeout}}: The robot exceeds maximum episode length.

\subsubsection{Unified Policy Training} To train the unified policy, we use the rewards listed in Tab~\ref{rewards}, and losses introduced in the main paper, where the hyperparameters are listed in Tab~\ref{loss_weight}.

\section{Humanoid Hiking Benchmark}\label{supp-benchmark}

This section provides: $(1)$ Qualitative comparisons of robot behaviors in response to varying trail challenges, demonstrating how different policy learning methodologies influence navigation and locomotion strategies tailored to humanoid tasks; $(2)$ Detailed quantitative results for each trail type between EP-H and RMA-B, offering insights into specific strengths and weaknesses of the approaches under distinct terrain and navigation conditions.

\noindent\textbf{Visualization.}
Fig~\ref{fig:sota-ablation} presents qualitative comparisons of LEGO-H with other benchmarked methods across five distinct trail examples, expanding on the key findings from Section~{\color{red}4.4} of the main paper. Additional insights include:$(1)$ without vision, RMA-B frequently fails to adapt to changing terrain properties (e.g., slope and surface friction) and falls over more often, as observed in Fig~\ref{fig:sota-ablation}(a)-(b). It also struggles to navigate obstacles effectively, often becoming stuck, as shown in Fig~\ref{fig:sota-ablation}(c). The higher MEV on Ditch and Hurdle, and lower trail completion on Forest in Tab~\ref{tab:hiking_benchmark_terrains} also demonstrate this. $(2)$ EP-H, which processes depth frames independently and applies brute-force cutoff for distant depth information, exhibits "circling" behaviors due to its inability to maintain scene continuity. This limitation hinders quick decision-making and recovery from self-induced distribution shifts, as demonstrated in Fig~\ref{fig:sota-ablation}(b), and results in inefficient navigation paths, as illustrated in Fig~\ref{fig:sota-ablation}(c). $(3)$ While leveraging vision, RMA-H lacks dynamic adaptability in navigation due to its separation of locomotion and navigation learning. This results in inefficient behaviors on trails requiring sharp turns or obstacle avoidance, as seen in Fig~\ref{fig:sota-ablation}(a)-(b). Additionally, its inefficient embodiment leads to unsafe detours, with trajectories that closely rub against obstacles, as highlighted in the zoomed-in trajectory in Fig~\ref{fig:sota-ablation}(c). $(4)$ The clean and safe-clearance trajectories of LEGO-H across all examples highlight the necessity and importance of integrative navigation and locomotion development through unified learning. 

\noindent\textbf{Insufficient Vision vs Blind.} Tab~\ref{tab:hiking_benchmark_terrains} show the comparison between EP-H and RMA-B. It indicates insufficient vision sometimes worse than blind vision.
\begin{figure}[h!]
    \centering
    \includegraphics[width=1\linewidth]{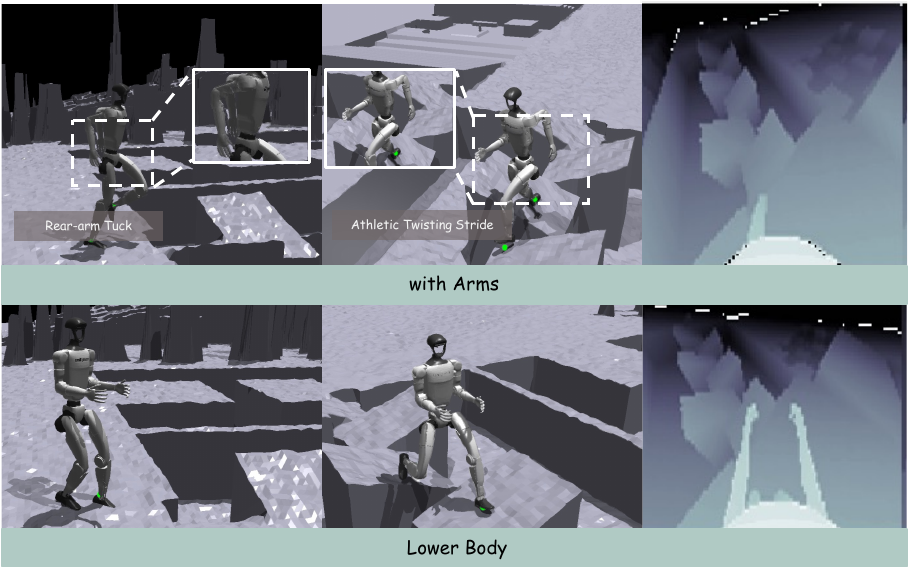}
\caption{{\textbf{Preliminary observations for future work on WBC.} G1  exhibits distinct motor behaviors over {\textit{with arms}} vs {\textit{only lower body}}. Besides, G1 emerges a rear-arm tuck posture while walking, likely to minimize arm interference with vision (see depth map).}}
    \label{fig:wbc-g1}
\end{figure}

\begin{figure*}[t!]
    \centering
    \includegraphics[width=1\linewidth]{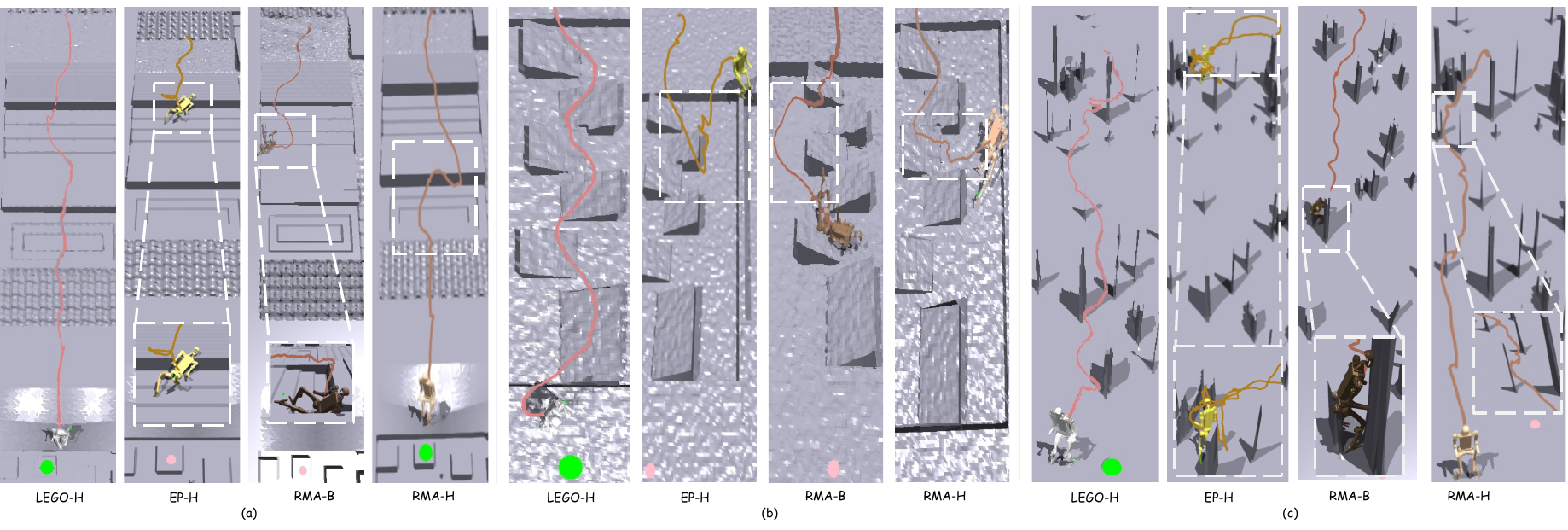}
    \caption{\textbf{Qualitative comparisons between LEGO-H and other benchmarked methods.}  The trajectories, visualized through dynamically updated colored lines, depict the robots' torso position as they traverse diverse trail environments. (a) illustrates the performance on a {\textit{RandomMix}} trail featuring unobstructed views with varied terrain types. (b) highlights the results on a {\textit{Ditch}} trail, where uneven terrain with slopes and gaps demands quick turns and agile leaps. (c) showcases the performance on a {\textit{Forest}} trail, where extensive obstacles of different sizes and heights block the robot’s view. The zoom-in regions highlight the issues of the robots. }

    \label{fig:sota-ablation}
\end{figure*}



\begin{table*}[h!]
\small
\caption{\textbf{EP-H vs RMA-B on each trail category.} This table employs a distinct protocol for fine-grained analysis: $256$ randomly initialized robots are evaluated for $30$ seconds {\textit{per}} trail category, spanning $25$ scenes ($5$ difficulty levels, each with $5$ variants). Results are averaged over $5$ runs to minimize random biases and ensure robustness.}

\label{tab:hiking_benchmark_terrains}
\resizebox{\textwidth}{!}{
    \begin{tabular}{c|c|c|c|c|c|c}
        \toprule
        \textbf{Methods} & \textbf{Success Rate (\%) $\uparrow$} & \textbf{Trail Completion (\%) $\uparrow$} & \textbf{Traverse Rate (\%) $\uparrow$} & \textbf{MEV (\%) $\downarrow$} & \textbf{TTF (s) $\uparrow$} & \textbf{Time-to-Reach (s) $\downarrow$} \\
        \midrule
        \multicolumn{7}{c}{\textbf{RandomMix}} \\
        \hline

        EP-H   & $16.98\pm0.85$ & $2.67\pm0.14$ & $70.88\pm1.41$ & $11.32\pm1.83$ & $3.33\pm0.13$ & $9.73\pm0.19$ \\

        RMA-B  & $30.99\pm0.95$ & $3.60\pm0.37$ & $76.74\pm1.13$ & $10.95\pm1.70$ & $4.14\pm0.20$ & $6.79\pm0.09$ \\
        \midrule
        \multicolumn{7}{c}{\textbf{Ditch}} \\
        \hline

        EP-H   & $16.12\pm0.66$ & $17.90\pm0.62$ & $55.75\pm0.58$ & $22.75\pm1.63$ & $3.50\pm0.08$ & $11.88\pm0.33$ \\

        RMA-B  & $32.80\pm1.56$ & $30.77\pm0.59$ & $63.49\pm1.42$ & $23.66\pm1.63$& $4.56\pm0.18$ & $6.37\pm0.37$ \\
        \midrule
        \multicolumn{7}{c}{\textbf{Hurdle}} \\
        \hline

        EP-H   & $46.54\pm2.64$ & $57.04\pm1.25$ & $68.95\pm1.79$ & $8.77\pm0.46$ & $6.44\pm0.21$ & $5.94\pm0.14$ \\

        RMA-B  & $83.04\pm0.27$ & $76.72\pm0.47$ & $83.04\pm1.17$ & $12.90\pm1.92$ & $9.24\pm0.28$ & $4.22\pm0.04$ \\
        \midrule
        \multicolumn{7}{c}{\textbf{Gap}} \\
        \hline

        EP-H   & $18.13\pm1.19$ & $32.26\pm0.48$ & $58.74\pm1.21$ & $31.84\pm2.00$ & $4.36\pm0.20$ & $12.15\pm0.34$ \\

        RMA-B  & $39.93\pm1.55$ & $44.27\pm1.06$ & $65.30\pm1.32$ & $24.10\pm2.09$ & $5.44\pm0.24$ & $7.99\pm0.17$ \\
        \midrule
        \multicolumn{7}{c}{\textbf{Forest}} \\
        \hline

        EP-H   & $63.29\pm1.50$ & $1.04\pm0.16$ & $82.61\pm1.18$ & $6.18\pm1.57$ & $8.96\pm0.49$ & $13.65\pm0.08$ \\

        RMA-B  & $64.81\pm2.43$ & $1.86\pm0.38$ & $81.59\pm3.20$ & $5.69\pm1.04$ & $10.18\pm0.89$ & $13.20\pm0.24$ \\
        \bottomrule
    \end{tabular}
}

\end{table*}

\section{Discussion}\label{future_work}

\noindent\textbf{Release and maintains.} All code/models/benchmarks will be publicly accessible and continuously updated to incorporate more robots/environments/models, aiming to establish a standard evaluation testbed for humanoid hiking research. We hope this project can lay the groundwork to promote future humanoid research from a new aspect.

\noindent\textbf{Future work.}  $(1)$ Kilometer-scale hiking. In this paper, we investigate humanoid robots on prototype trails to establish a baseline on the importance of integrative high-level navigation and low-level motor skills. However, real-world trails are considerably more complex, with long-distance traverse challenges. Future work could expand the framework to handle kilometer-scale trails, where sustained adaptability, energy efficiency, and long-term planning become crucial. $(2)$ Whole-body control for integrative navigation and locomotion skills. Expanding control across the entire body would enable a wider spectrum and adaptive behaviors, enhancing the robot's flexibility in complex, obstacle-rich environments. Our preliminary results suggest that while robots exhibit distinct motor styles based on physical constraints(Fig.~\ref{fig:wbc-g1}), {\textit{direct}} involvement of the upper body does not significantly impact performance. This opens opportunities for future work on exploring how coordinated whole-body strategies can enhance performance. $(3)$ Simulated environment upgrading. Our current simulated trails are primarily for foot contact; Future work could upgrade the simulated environment to better incorporate whole-body interactions, enabling a better testbed for future hiking studies.
$(4)$ Real-world deployment. In this paper, we conduct experiments on the simulator, enabling controlled benchmarking, rapid iteration, and reproducibility — {\textit{key prerequisites}} for real-world deployment. However, applying LEGO-H to real-world scenarios remains a vital next step toward closing the sim-to-real gap and realizing field-ready humanoid hikers.

\end{document}